\documentclass[journal]{IEEEtai}

\usepackage[colorlinks,urlcolor=blue,linkcolor=blue,citecolor=blue]{hyperref}

\usepackage{color,array}

\usepackage{graphicx}
\let\labelindent\relax
\usepackage{enumitem}
\usepackage{amsmath,amsfonts}
\usepackage{algorithmicx}
\usepackage{algorithm}
\usepackage[noend]{algpseudocode}
\usepackage{textcomp}
\usepackage{stfloats}
\usepackage{url}
\usepackage{verbatim}
\usepackage{cite}
\usepackage{adjustbox, makecell, multirow, multicol}
\usepackage[table]{xcolor}
\usepackage{color}
\usepackage{booktabs}

\DeclareMathOperator{\KL}{KL}

\DeclareMathOperator{\Softmax}{\sigma}

\DeclareMathOperator{\RL}{RL}
\DeclareMathOperator{\TL}{TL}
\DeclareMathOperator{\CL}{CL}
\DeclareMathOperator{\BoW}{BoW}
\DeclareMathOperator{\negt}{neg}
\DeclareMathOperator{\RBO}{RBO}
\DeclareMathOperator{\IRBO}{IRBO}
\DeclareMathOperator{\Encoder}{Encoder}
\renewcommand{\vec}[1]{\boldsymbol{\mathbf{#1}}}

\newcommand{\etal}{\textit{et al.}}

\begin{document}

\title{Evaluating Negative Sampling Approaches for Neural Topic Models}

\author{Suman Adhya,  Avishek Lahiri, Debarshi Kumar Sanyal, Partha Pratim Das
\thanks{``This work was supported in part by the Department of Science and Technology, Government of India under Grant CRG/2021/000803.'' }
\thanks{Suman Adhya, Avishek Lahiri, and Debarshi Kumar Sanyal are with the School of Mathematical and Computational Sciences, Indian Association for the Cultivation of Science, 2A \& 2B, Raja S C Mullick Road, Kolkata 700032, INDIA (e-mail: adhyasuman30@gmail.com, avisheklahiri2014@gmail.com, debarshi.sanyal@iacs.res.in).}
\thanks{Partha Pratim Das is with the Department of Computer Science, Ashoka University, Sonipat, Haryana 131029, INDIA, and Department of Computer Science and  Engineering, Indian Institute of Technology Kharagpur, West Bengal 721302, INDIA (e-mail: ppd@cse.iitkgp.ac.in).}
}


\maketitle

\begin{abstract}
Negative sampling has emerged as an effective technique that enables deep learning models to learn better representations by introducing the paradigm of ``learn-to-compare''. The goal of this approach is to add robustness to deep learning models to learn better representation by comparing the positive samples against the negative ones. Despite its numerous demonstrations in various areas of computer vision and natural language processing, a comprehensive study of the effect of negative sampling in an unsupervised domain like topic modeling has not been well explored. In this paper, we present a comprehensive analysis of the impact of different negative sampling strategies on neural topic models. We compare the performance of several popular neural topic models by incorporating a negative sampling technique in the decoder of variational autoencoder based neural topic models. Experiments on four publicly available datasets demonstrate that integrating negative sampling into topic models results in significant enhancements across multiple aspects, including improved topic coherence,  richer topic diversity, and more accurate document classification. Manual evaluations also indicate that the inclusion of negative sampling into neural topic models enhances the quality of the generated topics. These findings highlight the potential of negative sampling as a valuable tool for advancing the effectiveness of neural topic models.
\end{abstract}

\begin{IEEEImpStatement}
With the rapid advancement of technology, there has been a significant increase in the availability of text documents in digital format. Categorizing these documents based on their underlying content is crucial to facilitate easy access for users. However, manual labeling of these documents with their corresponding domain tags can be laborious and time-consuming due to the large volume of the corpus. Topic modeling techniques have emerged as a valuable tool in this context, as they can extract latent topics from a large corpus and label the documents with their dominant topics in an unsupervised manner. While traditional models pose computational challenges, neural topic models offer enhanced flexibility and scalability. Negative sampling-based models emphasize learning the document similarities and distinctions, thereby improving the quality of the learned topics. This paper conducts an empirical evaluation of several neural topic models based on negative sampling, implemented within a standardized framework to facilitate reproducibility. Thus, it will assist researchers and practitioners in identifying the most effective negative sampling strategies that may be employed for improving topic models.
\end{IEEEImpStatement}

\begin{IEEEkeywords}
Neural topic model, Variational autoencoder, Negative sampling, Contrastive learning, Topic coherence.
\end{IEEEkeywords}

\section{Introduction}
\IEEEPARstart{I}n topic modeling, latent variables referred to as topics are described as distributions over words. Given a corpus, a document is modeled as a random mixture over topics. The objective of topic modeling is to compute the posterior distribution of topics with the aim of reconstructing the given input document. However, determining the exact posterior is often impractical \cite{blei2017variational}. Consequently, various inferencing algorithms are employed to approximate the true posterior distribution. 
Nonetheless, these methods encounter a notable drawback due to their high computational costs and the necessity for re-derivation of the inference method when minor alterations occur in the modeling assumptions. To tackle these challenges, Miao \etal \cite{miao2016neural} introduced a neural topic model grounded in the Variational Autoencoder (VAE) framework. Recent advancements in this field have focused on incorporating diverse distributions on the document-topic distributions, integrating word embeddings, contextual information from documents, and utilizing available metadata from texts \cite{miao2017discovering, srivastava2017autoencoding, card2018neural, dieng2020topic, bianchi2020pre, bianchi2021cross, adhya2024ginopic}. 

Another approach to improve the performance of topic models is negative sampling: it introduces noise during the training process to enhance the model's robustness to out-of-distribution samples and mitigate the representation collapse problem \cite{xu2022negative, yang2024does}. Recently, we proposed a negative sampling-based method integrated with a contextualized neural topic model \cite{adhya2022neg}, that generates document-topic vectors as well as their perturbed forms, given a corpus, and uses a triplet loss so that the document reconstructed from the correct document-topic vector is similar to the input document but dissimilar to the one reconstructed from the perturbed vector. However, this negative sampling methodology is quite generic and can be integrated with other neural topic models. In this paper, we perform this integration and analyze the performance of seven high-performance neural topic models enhanced with the above negative sampling strategy\footnote{Code is available at \url{https://github.com/AdhyaSuman/Eval_NegTM}}. To make a comprehensive analysis, we compare the augmented models not only with their original versions but also with existing topic models that use contrastive learning. We observe that on most of the datasets that we experimented with, the performance, in terms of topic coherence and diversity, improves when a topic model is enhanced with our negative sampling strategy.

Our contributions can be summarized as follows:
\begin{itemize}
    \item We incorporate a negative sampling strategy into seven neural topic models, and assess its effectiveness and robustness. 
    \item We bring together a diverse set of neural topic models into a unified framework that categorizes the models based on the negative sampling strategy employed in them. 
    \item We present a comprehensive analysis of the above topic models assessing the influence of negative sampling through quantitative, qualitative, and task-specific evaluations. Additionally, we visualize the latent spaces generated by these models to assess their ability to disentangle the latent representations of documents. 
    \item Our assessments demonstrate that topic models incorporating negative sampling surpass those without this strategy for almost all datasets and experimental settings considered.  The visualizations also highlight a discernible enhancement in disentanglement ability when negative sampling is incorporated.
\end{itemize}

The remaining sections of the paper are organized as follows. In Section \ref{sec:Related_work}, we provide an overview of the evolution of neural topic modeling, highlighting works related to the scope of this paper.  Section \ref{sec:Taxonomy} presents a comprehensive description of existing negative sampling approaches, conducting a detailed analysis of their methodological components and associated objective functions. Section \ref{sec:datasets} contains the details on the datasets and model configurations used in evaluations. The empirical investigations are detailed in Section \ref{sec:results}, where we describe into the specifics of experimental configurations, evaluation methodologies, and the conclusions drawn from these empirical inquiries. Moving forward, Section \ref{sec:robustness} presents an evaluation of the impact of different vocabulary sizes on our proposed negative sampling approach, and how the modifications due to the incorporation of negative sampling strategy to the models affect the training time of the models. Additionally, in Section \ref{sec:LLM_based}, we provide a comparative analysis of our method with a recently proposed LLM-based topic model. The paper concludes in Section \ref{sec:conclusion}.

\section{Related Work} \label{sec:Related_work}
Topic modeling assumes that a document can be represented as a distribution over topics, where each topic is a distribution over the entire vocabulary. 
A popular approach for topic modeling is Latent Dirichlet Allocation (LDA) \cite{blei2003lda} in which the document-topic distribution is modeled as a Dirichlet distribution. 
However, determining the exact posterior distribution of the hidden variables in LDA is impractical \cite{blei2017variational}. Various approximate inference methods have been proposed to approximately determine these distribution parameters, such as Gibbs sampling \cite{gelfand1990mcmc}, Expectation-Maximization (EM) \cite{minka2002expectation}, and Variational Bayes inference \cite{blei2003lda}. Gibbs sampling, a Markov chain Monte Carlo algorithm, aids in statistical inferencing. Though it is a simple and effective technique, it has a high computational cost. On the other hand, the EM algorithm iteratively calculates the expected log-likelihood and optimizes parameter values. Variational Bayes (VB) constructs a lower bound for data likelihood and selects parameters maximizing this bound. Monte Carlo techniques provide \textit{\textit{numerical approximations}}, whereas VB offers an exact \textit{analytical} solution to the \textit{posterior approximation}. Thus, computing the posterior is quite challenging. To handle some of these issues, the Neural Variational Document Model (NVDM) \cite{miao2016neural} introduces the AutoEncoding Variational Bayes (AEVB) \cite{kingma2013auto} technique, commonly known as ``black-box inference'', into topic modeling. It is built with a VAE, which comprises an encoder and a decoder. In a high-level description, the encoder takes an input vector and projects it into a lower-dimensional latent space, while the decoder reconstructs the input vector from this latent representation. During training, this model seeks to maximize the tractable evidence lower bound to maximize the likelihood of the data distribution. Successive works in this direction produced models such as the Gaussian Softmax  (GSM) \cite{miao2017discovering}, Autoencoding Variational Inference For Topic Models (AVITM) \cite{srivastava2017autoencoding}, Sparse Contextual Hidden and Observed Language AutoencodeR (SCHOLAR) \cite{card2018neural}, Embedded Topic Model (ETM) \cite{dieng2020topic}, CombinedTM \cite{bianchi2020pre} and ZeroshotTM \cite{bianchi2021cross}. 

Negative sampling introduces a ``learn-to-compare'' paradigm \cite{chopra2005learning}, employing an effective strategy for learning rich representations. This technique gained popularity in NLP tasks, notably in the word2vec algorithm \cite{mikolov2013w2v}, to enhance training efficiency. In the domain of topic modeling, introducing negative samples in the VAE (encoder \cite{nguyen2021contrastive} and decoder \cite{wu2020nqtm, adhya2022neg}) and employing adversarial learning \cite{wang2018atm, wang2020bat, hu2020neural} have been shown to increase the quality of the generated topics. The negative sampling-based technique \cite{adhya2022neg} that we proposed only for the contextualized neural topic model \cite{bianchi2020pre} is a simple yet effective strategy, and we, therefore, integrate it with seven contemporary efficient neural topic models in this paper. We compare the enhanced topic models with their original versions (without negative sampling) and other related models proposed by researchers. Our previous work is thus significantly expanded in the current paper into a comprehensive evaluation of various negative sampling strategies in neural topic models. The OCTIS \cite{terragni2020octis} and ToModAPI \cite{lisena2020tomodapi} frameworks, proposed recently, lack consideration for topic models with negative sampling. This paper utilizes OCTIS to introduce negative sampling into neural topic models, creating an open-source framework for fair model comparisons. This study addresses the existing gap in the literature by offering a detailed analysis of negative sampling in neural topic models.

\section{Negative Sampling Strategies in Neural Topic Models} \label{sec:Taxonomy}
\begin{figure}
    \centering
    \includegraphics[width=\linewidth]{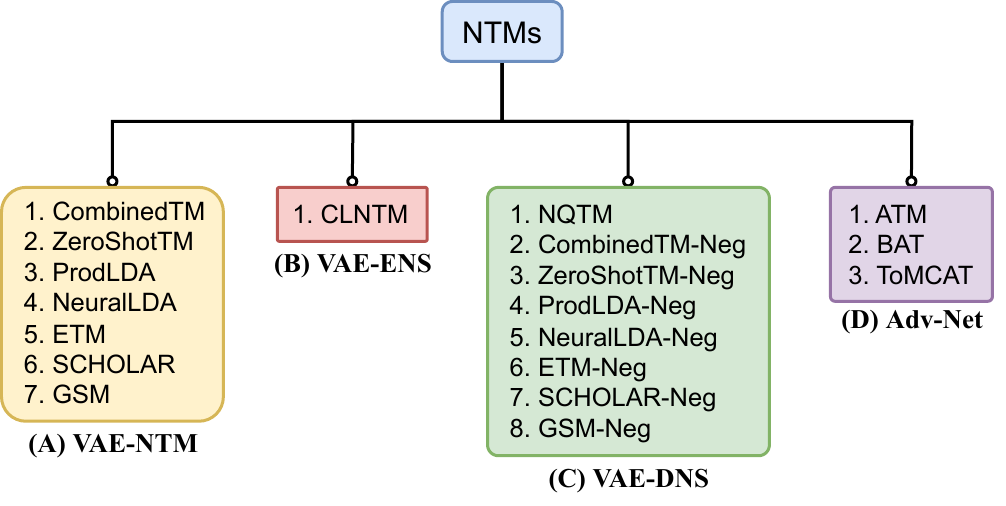}
    \caption{Topic model categories based on negative sampling techniques.}
    \label{fig:category}
\end{figure}

Negative sampling techniques can assist neural topic models to learn better latent representations. We previously introduced a negative sampling strategy \cite{adhya2022neg}, which involves perturbing the document-topic vector generated by the encoder and feeding it into the decoder. A triplet loss function ensures that the document reconstructed from the unperturbed topic vector is similar to the input document while being dissimilar to the document reconstructed from the perturbed vector. Thus, this technique effectively modifies the VAE decoder.

However, the literature on topic modeling includes instances where alternative forms of contrastive learning have been applied. For instance, negative document samples might be directly input to the encoder, or an adversarial network architecture could replace the VAE model. To facilitate a comparative examination, we present a taxonomy that encompasses these diverse VAE-based neural topic models. 
In particular, we categorize them into four distinct groups as follows:
\begin{enumerate}[label=(\Alph*)]
    \item VAE-based neural topic models \textit{without negative sampling} (\textbf{VAE-NTM});
    \item VAE-NTM with \textit{encoder negative sampling} (\textbf{VAE-ENS});
    \item VAE-NTM with \textit{decoder negative sampling} (\textbf{VAE-DNS});
    \item Adversarial network-based topic models (\textbf{Adv-Net});
\end{enumerate}

\begin{table}[!ht]
\caption{Notations and terminologies.\label{tab:Notations}}
    \centering
    \begin{adjustbox}{width=\linewidth}
    \begin{tabular}{c | l}
    \toprule
    \textbf{Symbol} & \textbf{Description} \\
    \midrule
    $V$ & Vocabulary size.\\
    $T$ & Number of topics.\\
    $\vec{x}$ & Document representation vector of size $V$.\\
    $(\vec{\mu}, \vec{\Sigma})$ & Mean and covariance, respectively,  in $T$-dimensional Gaussian.\\
    $\vec{z}$ & Latent representation of size $T$.\\
    $\sigma(\cdot)$ & $\texttt{Softmax}$ function.\\
    $\vec{\theta}$ & Document-topic distribution vector of size $T$.\\
    $\vec{\beta}$ & Topic-word matrix of dimension $T \times V$.\\
    $\hat{\vec{x}}$ & Reconstructed document representation of size $V$.\\
    \bottomrule
    \end{tabular}
    \end{adjustbox}
\end{table}

The topic models in each category are mentioned in Figure \ref{fig:category}. We will now describe each category in detail. Table \ref{tab:Notations} mentions the symbols that will be used in the description.

\subsection{VAE-NTM}
This category includes VAE-based neural topic models without negative sampling. Within this category, a general technique is AVITM \cite{srivastava2017autoencoding}, which incorporates a Laplace approximation to the Dirichlet prior \cite{mackay1998choice}. It includes models like ProdLDA and NeuralLDA \cite{srivastava2017autoencoding}. ProdLDA, by relaxing constraints on topic-word distributions, outperforms NeuralLDA. ETM \cite{dieng2020topic} uses word2vec embeddings for robustness, while SCHOLAR \cite{card2018neural} incorporates document metadata. The GSM model \cite{miao2017discovering} employs a softmax function on the document-topic vector.

In VAE-NTMs, the encoder takes a document representation vector ($\vec{x}$) as input. This vector can be either a normalized bag-of-words vector ($\vec{x} = \vec{x}_{\BoW}$) or a term frequency-inverse document frequency (tf-idf) vector ($\vec{x} = \vec{x}_{\operatorname{TF-IDF}}$). For CombinedTM \cite{bianchi2020pre}, a contextualized representation of the document ($\vec{x}_c$) is used alongside the bag-of-words representation ($\vec{x}_{\BoW}$). On the other hand, ZeroShotTM \cite{bianchi2021cross} relies solely on the contextualized representation of the document as input, allowing for training in one language and testing in another.

\begin{figure}
    \centering
    \includegraphics[width=\linewidth]{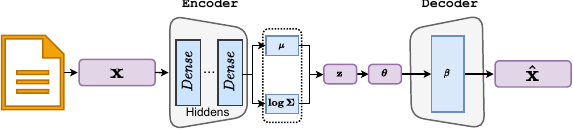}
    \caption{A general framework for VAE-based topic models.}
    \label{fig:VAE_NTMs}
\end{figure}

Given a Gaussian distribution as a prior and the input document representation vector $\vec{x}$, the encoder returns the variational posterior $q_{\vec{\phi}}(\vec{z} \vert \vec{x})$, where $\vec{\phi}$ represents the encoder weights. We assume the variational posterior to be the Gaussian distribution $\mathcal{N} \left(\vec{\mu},  \vec{\Sigma} \right)$, where $\vec{\mu}$ is the mean, and $\vec{\Sigma}$ represents the diagonal-covariance matrix. A latent representation $\vec{z}$ is stochastically sampled from the conditional distribution $q_{\vec{\phi}}(\vec{z} \vert \vec{x})$ using the reparameterization trick \cite{kingma2013auto}, given by $\vec{z} = \vec{\mu} + \vec{\Sigma}^{1/2} \odot \vec{\epsilon}$, where $\odot$ denotes the Frobenius inner product and $\vec{\epsilon} \sim \mathcal{N}(\vec{0}, \vec{I})$ is random noise. In the decoder, this latent representation $\vec{z}$ serves as the logits of a \texttt{softmax} function to generate the document-topic distribution $\vec{\theta}$ for reconstructing the input document, $\hat{\vec{x}} = \Softmax ( \vec{\beta}^{\top} {\vec{\theta}} )$, where $\vec{\beta}$ denotes the decoder weights. The loss function captures both the reconstruction error ($\mathcal{L}_{\RL}$) and the KL-divergence ($\mathcal{L}_{\KL}$) between the prior and posterior of the latent distribution:
\begin{align}
   \label{eq:VAE_loss}
    \mathcal{L} &= \mathcal{L}_{\RL} + \mathcal{L}_{\KL} \\
    &\equiv - \mathbb{E}_{\vec{z}
    \sim q_{\vec{\phi}}(\vec{z} \vert \vec{x})} [p_{\vec{\beta}}(\vec{x} \vert \vec{z})]  + \operatorname{D}_{\KL}\left(q_{\vec{\phi}}(\vec{z} \vert \vec{x})\| p(\vec{z})\right) \nonumber
\end{align}
A general framework of VAE-based neural topic models is illustrated in Figure \ref{fig:VAE_NTMs}. Since the above VAE-based architecture uses a Gaussian prior (unlike LDA), AVITM \cite{srivastava2017autoencoding} and models based on it have used a Laplace approximation to the Dirichlet.

\subsection{VAE-ENS}   
\begin{figure}
    \centering
    \includegraphics[width=\linewidth]{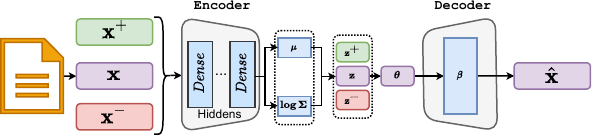}
    \caption{A general framework for the VAE-based topic models that uses negative sampling methodology on the encoder.}
    \label{fig:VAE_ENS}
\end{figure}

CLNTM \cite{nguyen2021contrastive} is the only VAE-based topic model that integrates a negative sampling technique within its encoder. The framework is outlined in Figure \ref{fig:VAE_ENS}. In CLNTM, given an input document representation $\vec{x}$, two additional samples are generated: a positive sample denoted by  $\vec{x}^{+}$ and a negative sample denoted by $\vec{x}^{-}$. These samples are converted to latent representations $\vec{z}^{+}$ and $\vec{z}^{-}$, respectively, by the encoder. To optimize the encoder and promote better separation between these latent representations, CLNTM employs the InfoNCE loss \cite{oord2019representation}:
\begin{align}
    \mathcal{L}_{\operatorname{InfoNCE}} = \mathbb{E}_{\vec{z}} \left[ - \log{ \left( \frac{\exp{(\vec{z} \cdot \vec{z}^{+})}}{\exp{(\vec{z} \cdot \vec{z}^{+})} + \eta \exp{(\vec{z} \cdot \vec{z}^{-})}} \right)}\right]
\end{align}
Here, $\eta$ serves as a hardness parameter of the negative samples. The overall objective function $\mathcal{L}_{\CL}$ for CLNTM is derived as: $\mathcal{L}_{\CL} = \mathcal{L} + \mathcal{L}_{\operatorname{InfoNCE}}$. In this equation, $\mathcal{L}$ represents the VAE loss function described in Eq. \eqref{eq:VAE_loss}. In their implementation\footnote{\url{https://github.com/nguyentthong/CLNTM}}, the authors set $\eta = 0.5$.

\subsection{VAE-DNS}
\begin{figure}
    \centering
    \includegraphics[width=\linewidth]{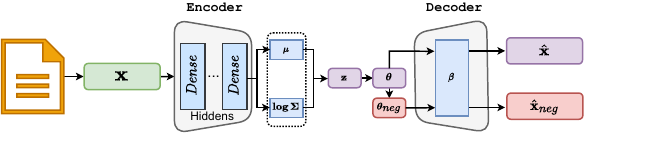}
    \caption{A general framework for the VAE-based topic models that uses negative sampling methodology on the decoder.}
    \label{fig:VAE_DNS}
\end{figure}

NQTM \cite{wu2020nqtm} and CombinedTM-Neg \cite{adhya2022neg} are VAE-based topic models incorporating negative sampling in the decoder. A general framework for VAE-DNS is shown in Figure \ref{fig:VAE_DNS}.

In NQTM, the goal is diverse topic generation for short texts. The process involves excluding top $t$ topics from the document-topic distribution and randomly drawing a negative sample, $\vec{z}_{\negt}$, from the remaining topics. A negative document, $\vec{x}_{\negt}$, is then generated from the top $M$ words of the distribution of the negative topic. This heuristic introduces a positive bias, aiding topic discovery, while the negative samples enhance learning signals. The objective function, $\mathcal{L}_{\CL}$, combines the VAE loss ($\mathcal{L}$), negative sampling loss ($\mathcal{L}_{\negt}$), and a proposed quantization loss ($\mathcal{L}_{\operatorname{quant}}$) to address topic collapsing in short texts as follows: $\mathcal{L}_{\CL} =  \mathcal{L} + \mathcal{L}_{\negt} + \mathcal{L}_{\operatorname{quant}}$.

The CombinedTM-Neg extends the contextualized topic model by incorporating a negative sampling technique within its decoder. This technique was proposed by us in \cite{adhya2022neg}. For each input document, a topic vector $\vec{\theta}$ is sampled, and a perturbed version $\tilde{\vec{\theta}}_{\negt}$ is generated from this vector. This perturbation involves setting the entries corresponding to the top $M$ topics to zero. Subsequently, $\tilde{\vec{\theta}}_{\negt}$ is normalized to obtain a probability vector $\vec{\theta}_{\negt}$. Thus, 
\begin{align}
    \vec{\theta}_{\negt} &= \frac{\tilde{\vec{\theta}}_{\negt}}{\sum_{i=1}^T \tilde{{\theta}}_{\negt}[i]} \nonumber \\
    \text{where } \tilde{{\theta}}_{\negt}[i]&= 
    \begin{cases}
      0 & \text{if topic $i$ is included in top $M$ topics,} \\
      {\theta}[i] & \text{otherwise}
    \end{cases}
\end{align}
It is assumed that $M < T$. The parameter $M$ is considered a hyperparameter to be tuned. Similar to the original topic vector $\vec{\theta}$, the perturbed topic vector $\vec{\theta}_{\negt}$ is then passed through the decoder network, which generates $\hat{\vec{x}}_{\negt} = \Softmax(\vec{\beta}^\top \vec{\theta}_{\negt})$, where $\vec{\beta}$ represents the topic-word matrix. The objective function is updated with the introduction of a new loss term known as triplet loss $\mathcal{L}_{\TL}$. In this loss, the anchor sample is $\hat{\vec{x}}$, which is the reconstructed output, the positive sample is $\vec{x}_{\BoW}$ (the original input document), and the negative sample is $\hat{\vec{x}}_{\negt}$.
    
\begin{align} \label{eq:triplet_loss}
    \mathcal{L}_{\TL} = \max\Big(\|\hat{\vec{x}} - \vec{x}_{\BoW}\|_2 - \|\hat{\vec{x}} - \hat{\vec{x}}_{\negt}\|_2 + m, 0 \Big) 
\end{align}
Here, $m$ represents the margin. The modified loss function to be minimized, denoted by $\mathcal{L}_{\CL}$, is a combination of the reconstruction loss $\mathcal{L}_{\RL}$, the KL-divergence loss $\mathcal{L}_{\KL}$, and the triplet loss $\mathcal{L}_{\TL}$, with the latter controlled by the hyperparameter $\lambda$.

\begin{align} \label{eq:loss2}
    \mathcal{L}_{\CL} &= \mathcal{L}_{\RL} + \mathcal{L}_{\KL} + \lambda \mathcal{L}_{\TL}
\end{align}

\begin{algorithm}
    \caption{CombinedTM-Neg \cite{adhya2022neg}}\label{alg:CombinedTM-Neg}
    \begin{algorithmic}[1]
    \While{(not converged)}
        \State $ \big( \vec{\mu}, \vec{\Sigma} \big) \leftarrow \Encoder(\vec{x})$
        
        \State $ \vec{z} \leftarrow \vec{\mu} + \vec{\Sigma}^{1/2} \odot \vec{\epsilon}$ \algorithmiccomment{$\vec{\epsilon} \sim \mathcal{N}(\vec{0}, \vec{I})$}
        
        \State $ \vec{\theta} \leftarrow \sigma \big(\vec{z}\big)$ \algorithmiccomment{$ \sigma(\cdot)$ is softmax function.}
        
        \State $\tilde{\vec{\theta}}_{\negt} \xleftarrow{\text{Perturbation}} \vec{\theta}$ 
        
        \State $\vec{\theta}_{\negt} \xleftarrow{\text{Normalization}} \tilde{\vec{\theta}}_{\negt}$ \algorithmiccomment{$\vec{\theta}_{\negt} = \tilde{\vec{\theta}}_{\negt}/\sum_{i=1}^T \tilde{{\theta}}_{\negt}[i]$}
        
        \State $\hat{\vec{x}} \leftarrow \Softmax ( \vec{\beta}^{\top} {\vec{\theta}} )$ \algorithmiccomment{Recon. of the original doc.}
        
        \State $\hat{\vec{x}}_{\negt} \leftarrow \Softmax ( \vec{\beta}^{\top} {\vec{\theta}}_{\negt})$ \algorithmiccomment{Recon. of a negative doc.}

        \State Compute: $\mathcal{L}_{\TL}, \mathcal{L}_{\RL}$ and $\mathcal{L}_{\KL}$
        
        \State $\mathcal{L}_{\CL} \xleftarrow{\text{Total loss}} \left(\mathcal{L}_{\RL} + \mathcal{L}_{\KL}\right) + \lambda \mathcal{L}_{\TL}$ 
        \State Update weights to minimize $\mathcal{L}_{\CL}$.
    \EndWhile
    \end{algorithmic}
\end{algorithm}

The complete methodology is described in Algorithm \ref{alg:CombinedTM-Neg}, and it is trained end-to-end using the Adam optimizer and backpropagation. Due to its inherent simplicity, we have incorporated it in many other topic models within the VAE-NTM category, namely, ProdLDA \cite{srivastava2017autoencoding}, NeuralLDA \cite{srivastava2017autoencoding}, ETM \cite{dieng2020topic}, ZeroShotTM \cite{bianchi2021cross}, SCHOLAR \cite{card2018neural}, and GSM \cite{miao2017discovering}.

\subsection{Adv-Net}
ATM \cite{wang2018atm}, BAT \cite{wang2020bat}, and ToMCAT \cite{hu2020neural} are GAN-based topic models employing negative sampling. ATM is the pioneer in this category. It has a document sampling module to get the tf-idf representations of input documents, a generator network ($G$) to generate fake documents from a Dirichlet prior, and a discriminator network ($D$) to distinguish between fake and real documents. The objective function, $\mathcal{L}_{\CL}$, includes discriminator loss ($\mathcal{L}_d$) and gradient penalty ($\mathcal{L}_{gp}$), with $\zeta$ adjusting the importance of the penalty. The overall objective function of ATM is, $\mathcal{L}_{\CL} = \mathcal{L}_d + \zeta \mathcal{L}_{gp}$.
However, ATM cannot infer document-topic distributions. BAT addresses this limitation by introducing an extra \textit{encoder network} ($E$) to infer document-topic distributions. The encoder takes the weighted tf-idf document representation $\vec{d}_r$ (real) as input and transforms it into the corresponding topic distribution $\vec{\theta}_r$ (real). On the other hand, the generator takes a random topic distribution $\vec{\theta}_f$ (fake) drawn from a Dirichlet prior and generates a fake word distribution $\vec{d}_f$. The discriminator receives a real distribution pair $\vec{p}_r = [\vec{\theta}_r ; \vec{d}_r]$ and a fake distribution pair $\vec{p}_f = [\vec{\theta}_f ; \vec{d}_f]$ as input and discriminates between them. The optimization objective for BAT is to minimize the Wasserstein distance between the generated joint distribution $\mathbb{P}_f$ and the real joint distribution $\mathbb{P}_r$ as follows:
\begin{align}
    \mathcal{L}_{\CL} = \mathbb{E}_{\vec{p_f} \sim \mathbb{P}_f} [D(\vec{p}_f)] - \mathbb{E}_{\vec{p_r} \sim \mathbb{P}_r} [D(\vec{p}_r)]
\end{align}
where $D(\cdot)$ denotes the output signal of the discriminator, with a higher value indicating a greater likelihood of considering the input as a real distribution pair, and vice-versa.

The third adversarial topic model is ToMCAT \cite{hu2020neural} which is based on the cycle-consistent adversarial training. As implementations for ATM and ToMCAT are not publicly available, we could not include them in our evaluation.

\section{Configuration of Datasets and Models}\label{sec:datasets}
We have used four publicly available datasets in our experiments. The \textbf{20NewsGroups} (\textbf{20NG}) \cite{terragni2020octis} dataset, featuring 16,309 newsgroup documents, is evenly distributed across 20 distinct newsgroups. The \textbf{GoogleNews} (\textbf{GN}) \cite{Qiang2020Short} collection, derived from the Google News website in November 2013, comprises 11,109 news articles, titles, and snippets. The \textbf{M10} \cite{Pan2016M10} corpus from CiteSeer$^\text{X}$ consists of 8,355 scientific publications spanning 10 research areas. Lastly, we incorporated a subset of 24,774 English documents from the \textbf{Wiki40B} \cite{Wiki40B} dataset, a comprehensive Wikipedia text collection. Detailed statistics for these datasets are available in Table \ref{tab:data_stat}. From the table, it can be observed that Wiki40B and 20NG have larger document sizes than those of GN and M10.

\begin{table}[!ht]
\caption{Statistics of the used datasets.\label{tab:data_stat}}
    \centering
    \begin{adjustbox}{width=1\linewidth}
    \begin{tabular}{l|c|c|c|c}
    \toprule
     \textbf{Dataset} & \textbf{Type} &\textbf{\#Docs} & \textbf{Avg. doc. len.} &  \textbf{Vocab. Size}\\ \midrule
    \textbf{20NG} & 20 different newsgroups & 16309 & 48.02 & 1612\\ 
    \textbf{GN} & Google news & 10920 & 5.35 & 2000\\ 
    \textbf{M10} & Scientific publications & 8355 & 5.91 & 1696\\ 
    \textbf{Wiki40B} & Wikipedia text & 24774 & 541.08 & 2000\\ 
    \bottomrule
    \end{tabular}
    \end{adjustbox}
\end{table}

In our experiments, we employed OCTIS \cite{terragni2020octis}, an integrated framework designed for comparing and evaluating topic models. Within the OCTIS framework, we integrated the implementations of the following models, as publicly provided by the authors of the respective papers: SCHOLAR \cite{card2018neural}, GSM \cite{miao2017discovering}, CLNTM \cite{nguyen2021contrastive}, NQTM \cite{wu2020nqtm}, and BAT \cite{wang2020bat}. For CombinedTM, ZeroShotTM, ProdLDA, NeuralLDA, and their negative sampling counterparts, a fully-connected feed-forward neural network (FFNN) with two hidden layers, each containing 100 neurons, was employed in the encoder. The decoder utilized a single-layer FFNN. To acquire contextualized embeddings of input documents for CombinedTM and ZeroShotTM, we utilized the \texttt{paraphrase-distilroberta-base-v2}\footnote{\url{https://huggingface.co/sentence-transformers/paraphrase-distilroberta-base-v2}}, an SBERT model \cite{reimers2019sentence}.

The models incorporating decoder negative sampling, as proposed in \cite{adhya2022neg}, utilized a triplet loss as outlined in Eq. \eqref{eq:triplet_loss}. The margin value $m$ in this equation was set to 1, following the default setting in PyTorch\footnote{\url{https://pytorch.org/docs/stable/generated/torch.nn.TripletMarginLoss.html}}. Hyperparameters $M$ and $\lambda$ were optimized using OCTIS's Bayesian optimization framework to maximize NPMI,  where $M \in \{1,2,3\}$ and $\lambda \in (0,1]$. Other hyperparameters were maintained at their default values.

For all datasets, we constrained the vocabulary to the top 2000 most common words in the corpus. However, in some of the datasets, the number of unique words is lower. Experiments for each topic model were conducted across various topic counts in the set $\{10, 20, 30, 40, 50, 60, 90, 120\}$.

Our experiments were run on a workstation with 6-core Intel\textsuperscript{\textregistered} Xeon\textsuperscript{\textregistered} W-1350 CPU, 16 GB RAM, NVIDIA RTX A4000 GPU, and Ubuntu 22.04.

\section{Results and Discussions}\label{sec:results}
\subsection{Quantitative Evaluation}
\begin{table*}[!ht]
\caption{Comparing topic models for various datasets. For each metric, we first calculate the median value across all runs for a given topic count, then average the medians over the topic counts. \textbf{Bold} highlights the highest value of a metric between a model with added decoder negative sampling and its vanilla counterparts. $\dag$ indicates the highest value of a metric over all models for a given dataset. \textbf{Bold} highlights the highest value of a metric between a model with added decoder negative sampling and its vanilla counterparts. $\dag$ indicates the highest value of a metric over all models for a given dataset. \label{tab:QuantACcompare}}
    \centering
    \begin{adjustbox}{width=1\linewidth}
    \begin{tabular}{ l | ccc  | ccc | ccc | ccc }
    \toprule
     \multirow{2}{*}{\textbf{Model}} & \multicolumn{3}{c|}{\textbf{20NG}} & \multicolumn{3}{c|}{\textbf{GN}} & \multicolumn{3}{c|}{\textbf{M10}} & \multicolumn{3}{c}{\textbf{Wiki40B}} \\ 
     
     & \textbf{NPMI} & \textbf{CV} & \textbf{IRBO} & \textbf{NPMI} & \textbf{CV} & \textbf{IRBO} & \textbf{NPMI} & \textbf{CV} & \textbf{IRBO} & \textbf{NPMI} & \textbf{CV} & \textbf{IRBO}\\ \midrule
      \textbf{CombinedTM} \cite{bianchi2020pre} & 0.093 & 0.627 & 0.990 & 0.081 & 0.485 & 0.995 & 0.048 & \textbf{0.466} & 0.980 & 0.133 & 0.647 & 0.988 \\
     
     \textbf{CombinedTM-Neg} \cite{adhya2022neg} & \textbf{0.121} & \textbf{0.648} & \textbf{0.991} & \textbf{0.142} & \textbf{0.530} & \textbf{0.998} & \textbf{0.052}$^\dag$ & 0.462 & \textbf{0.986} & \textbf{0.139} & \textbf{0.661} & \textbf{0.990} \\ \midrule

      \textbf{ZeroShotTM} \cite{bianchi2021cross} & 0.115 & 0.643 & 0.992 & 0.115 & 0.520 & \textbf{0.999} & 0.029 & \textbf{0.443} & \textbf{0.991} & 0.137 & 0.658 & 0.989\\
     
     \textbf{ZeroShotTM-Neg} & \textbf{0.126} & \textbf{0.654} & \textbf{0.993} & \textbf{0.143} & \textbf{0.541}$^\dag$ & \textbf{0.999} & \textbf{0.037} & 0.442 & \textbf{0.991} & \textbf{0.144} & \textbf{0.669} & \textbf{0.990} \\ \midrule

      \textbf{ProdLDA} \cite{srivastava2017autoencoding} & 0.080 & 0.609 & 0.990 & 0.056 & 0.471 & 0.996 & 0.025 & \textbf{0.448} & 0.983 & 0.126 & 0.635 & 0.983 \\
     
    \textbf{ProdLDA-Neg} & \textbf{0.105} & \textbf{0.623} & \textbf{0.992} & \textbf{0.144}$^\dag$ & \textbf{0.538} & \textbf{0.998} & \textbf{0.032} & 0.447 & \textbf{0.988} & \textbf{0.134} & \textbf{0.652} & \textbf{0.986} \\ \midrule

      \textbf{NeuralLDA} \cite{srivastava2017autoencoding} & 0.058 & 0.526 & \textbf{0.998}$^\dag$ & -0.120 & 0.407 & \textbf{1.000}$^\dag$ & -0.207 & 0.387 & \textbf{0.999}$^\dag$ & 0.074 & 0.531 & 0.997 \\
     
     \textbf{NeuralLDA-Neg} & \textbf{0.066} & \textbf{0.536} & \textbf{0.998}$^\dag$ & \textbf{-0.090} & \textbf{0.413} & \textbf{1.000}$^\dag$ & \textbf{-0.189} & \textbf{0.392} & \textbf{0.999}$^\dag$ & \textbf{0.079} & \textbf{0.542} & \textbf{0.998}$^\dag$ \\ \midrule

      \textbf{ETM} \cite{dieng2020topic} & 0.049 & 0.528 & 0.819 & -0.263 & \textbf{0.414} & 0.627 & -0.056 & 0.345 & \textbf{0.502} & 0.080 & 0.554 & 0.952 \\
     
     \textbf{ETM-Neg} & \textbf{0.054} & \textbf{0.540} & \textbf{0.826} & \textbf{-0.224} & 0.410 & \textbf{0.638} & \textbf{-0.032} & \textbf{0.352} & 0.485 & \textbf{0.087} & \textbf{0.571} & \textbf{0.957} \\

     \textbf{ETM-Encoder-Neg} & 0.045 & 0.521 & 0.816 & -0.255 & 0.409 & 0.664 & -0.058 & 0.344 & 0.476 & 0.064 & 0.526 & 0.940 \\ \midrule
     
      \textbf{SCHOLAR} \cite{card2018neural} & 0.172 & 0.760 & \textbf{0.995} & -0.201 & \textbf{0.529} & 0.973 & -0.090 & \textbf{0.464} & \textbf{0.800} & 0.175 & 0.756 & \textbf{0.995} \\
     
      \textbf{SCHOLAR-Neg} & \textbf{0.180}$^\dag$ & \textbf{0.764}$^\dag$ & \textbf{0.995} & \textbf{-0.150} & \textbf{0.529} & \textbf{0.982} & \textbf{-0.017} & {0.460} & {0.779} & \textbf{0.181}$^\dag$ & \textbf{0.760}$^\dag$ & \textbf{0.995} \\ \midrule

      \textbf{GSM} \cite{miao2017discovering} & 0.089 & 0.560 & \textbf{0.979} & -0.018 & \textbf{0.482} & \textbf{0.951} & -0.007 & 0.368 & \textbf{0.862} & 0.080 & 0.505 & \textbf{0.989} \\
     
     \textbf{GSM-Neg} & \textbf{0.095} & \textbf{0.569} & 0.977 & \textbf{0.012} & 0.475 & 0.947 & \textbf{0.021} & \textbf{0.378} & 0.841 & \textbf{0.087} & \textbf{0.520} & \textbf{0.989} \\ \midrule
     
     \textbf{CLNTM} \cite{nguyen2021contrastive} & 0.106 & 0.691 & 0.992 & -0.362 & 0.491 & 0.969 & -0.381 & 0.473 & 0.923 & 0.088 & 0.630 & 0.990 \\ 
     
     \textbf{NQTM} \cite{wu2020nqtm} & -0.021 & 0.433 & 0.987 & -0.452 & 0.540 & 0.991 & -0.402 & 0.490$^\dag$ & 0.987 & -0.013 & 0.336 & 0.974 \\
     
     \textbf{BAT} \cite{wang2020bat} & -0.001 & 0.419 & 0.929 & -0.276 & 0.466 & 0.931 & -0.364 & 0.453 & 0.908 & 0.101 & 0.581 & 0.977 \\ \bottomrule
    \end{tabular}
    \end{adjustbox}
\end{table*}

\begin{figure*}
    \centering
    \includegraphics[width=\textwidth]{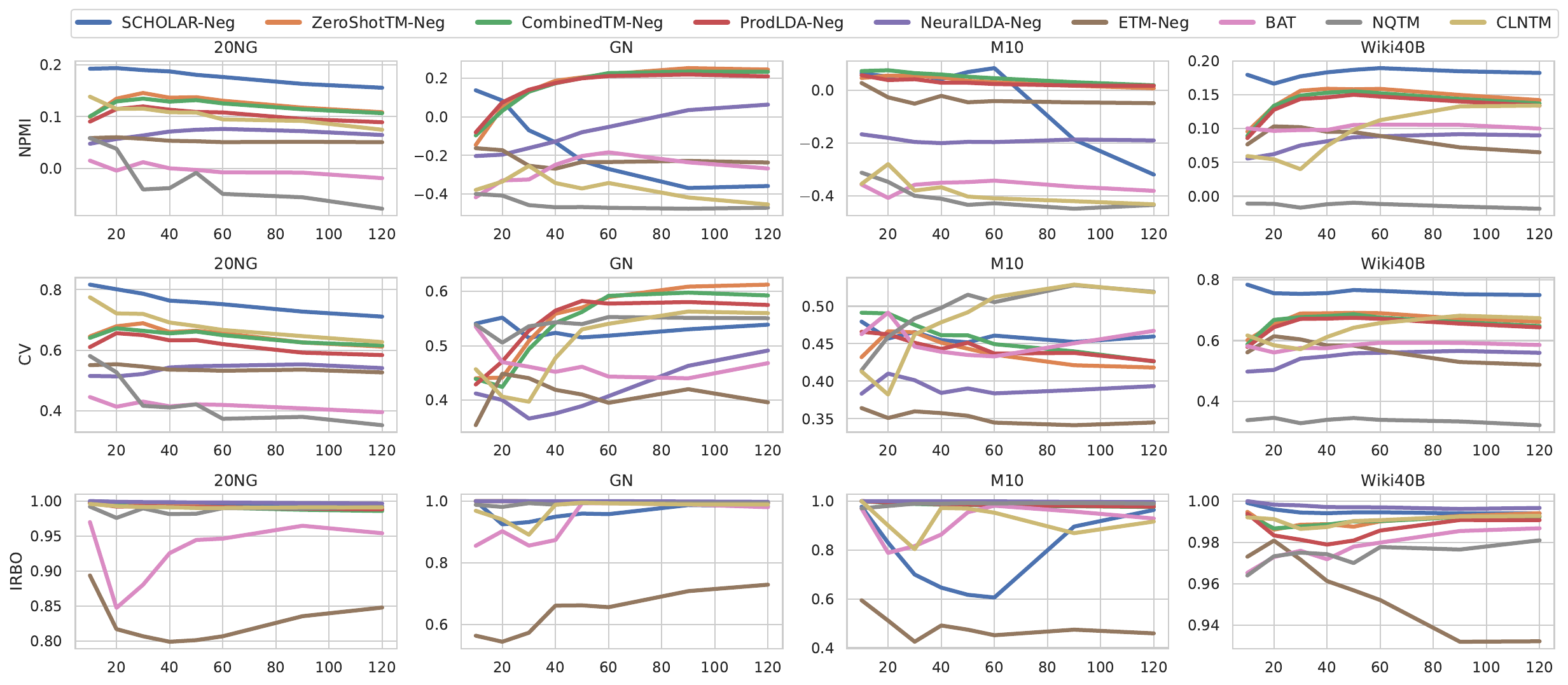}
    \caption{The variation of topic coherence (NPMI and CV) and topic diversity (IRBO) with topic count are shown for different topic models on four datasets. The ordinate value of each data point reports the median over five independent runs.}
    \label{fig:compareTM}
\end{figure*}

In order to quantitatively evaluate the performance of the above topic models, we conducted a comprehensive experimental analysis for four different datasets.

\subsubsection{Experimental setup}
We have varied the topic count, denoted by $T$, over a range of values, namely $\{10, 20, 30, 40, 50, 60, 90, 120\}$. In order to ensure robustness, for each topic count, we have run the models $5$ times with different random seeds and then taken the median value of the metric. Finally, we have averaged the values of the metric over the different topic counts. This approach effectively mitigates the influence of outliers and fluctuations, providing a more reliable assessment of the model's performance. 

\subsubsection{Evaluation metrics}
Topic models are commonly evaluated using two key metrics: coherence score and topic diversity. These metrics provide insights into the quality and distinctiveness of the generated topics.

\textbf{Coherence measure:} It is used to compute the relatedness of the prominent words within a topic \cite{hoyle2021automated}. We use \textbf{Normalized Pointwise Mutual Information} (\textbf{NPMI}) \cite{lau2014machine} and \textbf{Coherence Value} (\textbf{CV}) \cite{roder2015exploring} to measure topic coherence. NPMI is widely adopted as a replacement for human judgment of topic coherence. Some researchers also use CV (but CV has some \href{https://palmetto.demos.dice-research.org/}{known issues}).
NPMI for $T$ topics each of which contains $t$ top words can be calculated as:
\begin{align}
    \operatorname{NPMI} = \frac{1}{T} \sum_{k=1}^T \frac{1}{\binom{t}{2}}\sum_{i=1}^t \sum_{j=i+1}^t \frac{\log{\frac{p(w_i^{(k)}, w_j^{(k)}) + \epsilon}{p(w_i^{(k)})p(w_j^{(k)})}}}{- \log{\left(p(w_i^{(k)}, w_j^{(k)})+\epsilon\right)}}
\end{align}
where $p(w_i^{(k)}, w_j^{(k)})$ is the probability of the co-occurrence of the words $w_i^{(k)}$ and $w_j^{(k)}$ in a boolean sliding window in topic $k$, and $p(w_i^{(k)})$ and $p(w_j^{(k)})$ represent the probability of the occurrence of the corresponding words in topic $k$. $\epsilon$ is a small positive constant that is used to avoid zero in the $\log(\cdot)$ function. NPMI lies in $[-1, \: +1]$ where $-1$ indicates the words never co-occur and $+1$ indicates they always co-occur. CV combines cosine similarity with  NPMI and boolean sliding window \cite{roder2015exploring, krasnashchok2018improving}.

\textbf{Topic diversity measure:} We measure the diversity of topics using \textbf{Inversed Rank-Biased Overlap} (\textbf{IRBO}) \cite{bianchi2020pre}. It assigns a value of $0$ to identical topics and $1$ to completely dissimilar topics, based on their word lists. Suppose we are given a collection of $T$ topics where for each topic the corresponding word list places higher-ranked words at the beginning. Then, the IRBO score of the topics is defined as:
\begin{align}
    \IRBO = 1 - \frac{\sum_{i=2}^T \sum_{j=1}^{i-1} \RBO(l_i, l_j)}{n} 
\end{align}
where $n = {\binom{T}{2}}$ is the number of pairs of lists, and $\RBO(l_i, l_j)$ \cite{webber2010similarity} denotes the standard Rank-Biased Overlap between two ranked lists $l_i$ and $l_j$. IRBO gives lower scores for overlapping words at higher ranks. Higher values of NPMI, CV, and IRBO are better than lower values. 

\subsubsection{Results}
In Table \ref{tab:QuantACcompare}, we present coherence values (NPMI and CV) and topic diversity (IRBO) across all topic models for all the datasets. The results indicate that \textit{models incorporating decoder negative sampling consistently generate topics with higher NPMI coherence scores} compared to their original versions that do not utilize negative sampling and the other models that use encoder negative sampling (CLNTM) or adversarial learning (BAT). For example, for 20NG, the average NPMI score is 0.172 for SCHOLAR and 0.180 for SCHOLAR-Neg; 0.115 for ZeroShotTM, 0.126 for ZeroShotTM-Neg; 0.106 for CLNTM; -0.021 for NQTM; and -0.001 for BAT. 
Regarding CV coherence scores, again the decoder negative sampling models generally outperform their counterparts without negative sampling. The IRBO score does not change appreciably after negative sampling is added. For example, both the SCHOLAR and SCHOLAR-Neg models achieve an IRBO score of 0.995 on the 20NG dataset. In summary, we observe an improvement in coherence scores without any negative impact in topic diversity. 

Let us now take a dataset-oriented view of Table \ref{tab:QuantACcompare}. For 20NG and Wiki40B, we observe that SCHOLAR-Neg outperforms all other models in coherence scores (NPMI and CV). For GN, ProdLDA-Neg achieves the highest NPMI (0.144), while ZeroShotTM-Neg attains the highest CV score (0.541). A closer look reveals that on the GN dataset, the three models CombinedTM-Neg, ZeroShotTM-Neg, and ProdLDA-Neg produce very similar scores for NPMI (0.142 to 0.144) and CV (0.530 to 0.541), and all of them are considerably better compared to their respective original versions which do not utilize negative samples.
For the M10 dataset, CombinedTM-Neg exhibits the highest NPMI while NQTM demonstrates the best CV score. Consequently, across all four datasets, methods utilizing negative sampling achieve the highest topic coherence. It is interesting to note that, for the GN and M10 corpora, which contain shorter documents compared to those in 20NG and Wiki40B, NQTM secures the second-highest and highest CV scores, respectively. This is probably because NQTM has been optimized for short documents \cite{wu2020nqtm}. We also observe that CV and NPMI exhibit a high correlation for 20NG and Wiki40B but not for GN and M10. This may be explained by noting that the two coherence metrics are known to show high correlation only if the sliding window used for CV calculation is large \cite{roder2015exploring} but the latter two datasets contain very short documents effectively precluding the use of a large sliding window.  In terms of topic diversity, NeuralLDA-Neg consistently performs the best for all datasets.

Figure \ref{fig:compareTM} illustrates the variation of NPMI, CV, and IRBO scores with topic count for each of the four datasets.
The figure indicates that the SCHOLAR-Neg model consistently generates the most coherent topics (NPMI and CV) for the 20NG and Wiki40B datasets. For the GN and M10 datasets, characterized by relatively shorter document lengths, CombinedTM-Neg, ZeroShotTM-Neg, and ProdLDA-Neg exhibit similar performance trends for NPMI and CV across all topic counts. This similarity may be attributed to the absence of major architectural differences between these three models, as all of them are built upon the ProdLDA architecture. Clearly, in terms of NPMI, these models outperform other models on GN and M10. The IRBO plots reveal that NeuralLDA-Neg consistently achieves the highest IRBO scores. Additionally, apart from ETM, most of the models display comparable IRBO scores across varying topic counts. ETM-Neg consistently generates significantly lower IRBO values, indicating a tendency towards repetitive topics. To investigate whether the encoder negative sampling strategy can improve ETM, we integrated the algorithm proposed by Nguyen \etal \cite{nguyen2021contrastive} into ETM, resulting in a modified version called ETM-Encoder-Neg. But it failed to enhance the performance. In fact, we observed a further decrease in performance in terms of both coherence and diversity metrics across all datasets except GN. 

\subsection{Qualitative Evaluation}
Several studies (e.g., \cite{hoyle2021automated}) have indicated that automated metrics may not always align well with human evaluations of topics. As a result, we conduct manual assessments to evaluate the quality of topics generated by the models.

\subsubsection{Experimental setup} 
To qualitatively assess model performance, we have used the Wiki40B dataset with a topic count of 120. Following the methodology proposed by Adhya \etal \cite{adhya2023neural}, we compared pairwise similar topics generated by different models. Suppose lists $P$ and $Q$  represent topics from two topic models respectively. We construct a similarity matrix with Rank-biased Overlap (RBO) scores, indicating the similarity between the topics. RBO scores range from 0 (no overlap) to 1 (complete overlap). We then select topic pairs with the largest similarity scores while ensuring each pair is unique, that is, if two selected topic pairs are $\big(P[i_1], Q[j_1]\big)$ and $\big(P[i_2], Q[j_2]\big)$, then $\left(i_1 \neq i_2 \land j_1 \neq j_2\right)$. Given the top 10 words in a topic, we look for \textit{odd} words, that is, words that do not appear to fit with the other words in the topic. A large proportion of \textit{odd words} in a topic reflects poorly on the quality of the topic.

\begin{table*}[!htbp]
\caption{Aligned topics from vanilla VAE-NTM and its decoder-negative sampling counterpart, trained on Wiki40B for $T=120$ topics. Closely related words in each topic are highlighted in \textbf{bold}. \label{tab:Qualitative}}
\centering
\begin{adjustbox}{width=.97\linewidth}
  \begin{tabular}{ l | l } \toprule
    \textbf{Model} & \multicolumn{1}{c}{\textbf{Topics}} \\ \midrule
     \multirow{3}{*}{\textbf{CombinedTM}\cite{bianchi2020pre}} & \textbf{coin}, refer, \textbf{issue}, letter, \textbf{currency}, etymology, commonly, word, adopt, flag \\
     & \textbf{missile}, \textbf{orbit}, \textbf{launch}, \textbf{satellite}, \textbf{aircraft}, \textbf{mission}, \textbf{flight}, \textbf{earth}, \textbf{space}, successfully \\ 
     & \textbf{work}, \textbf{write}, die, \textbf{painting}, \textbf{publish}, \textbf{prize}, move, paris, father, professor \\ \cline{2-2}
    \multirow{3}{*}{\textbf{CombinedTM-Neg}\cite{adhya2022neg}} & \textbf{currency}, \textbf{bank}, \textbf{exchange}, \textbf{money}, \textbf{rate}, \textbf{credit}, \textbf{coin}, \textbf{crisis}, \textbf{financial}, \textbf{reserve} \\
    & \textbf{orbit}, \textbf{missile}, \textbf{launch}, \textbf{mission}, \textbf{satellite}, \textbf{aircraft}, \textbf{flight}, \textbf{moon}, \textbf{earth}, successfully \\
    & career, \textbf{writer}, \textbf{publish}, \textbf{award}, \textbf{hindi}, \textbf{poet}, \textbf{literary}, \textbf{literature}, \textbf{poetry}, bear \\ \midrule 
     \multirow{3}{*}{\textbf{ZeroShotTM}\cite{bianchi2021cross}} & \textbf{disease}, \textbf{heart}, \textbf{symptom}, \textbf{risk}, \textbf{pain}, \textbf{occur}, recommend, people, include, increase \\
     & \textbf{corruption}, \textbf{defence}, \textbf{missile}, \textbf{nuclear}, company, india, \textbf{security}, \textbf{intelligence}, indian, \textbf{investigation} \\
     & \textbf{write}, \textbf{philosophy}, \textbf{work}, \textbf{thought}, \textbf{read}, \textbf{publish}, \textbf{writing}, later, idea, friend \\ \cline{2-2}
     \multirow{3}{*}{\textbf{ZeroShotTM-Neg}} & \textbf{infection}, \textbf{virus}, \textbf{disease}, \textbf{transmission}, \textbf{symptom}, \textbf{cause}, \textbf{transmit}, \textbf{host}, \textbf{protein}, \textbf{spread} \\
     & \textbf{nuclear}, \textbf{nations}, \textbf{bomb}, \textbf{weapon}, \textbf{war}, \textbf{soviet}, \textbf{united}, \textbf{peace}, \textbf{global}, \textbf{radiation} \\
     & \textbf{writer}, \textbf{publish}, \textbf{novel}, \textbf{literature}, \textbf{write}, \textbf{literary}, \textbf{magazine}, \textbf{poet}, \textbf{poem}, \textbf{story} \\ \midrule 
     \multirow{3}{*}{\textbf{ProdLDA}\cite{srivastava2017autoencoding}} & \textbf{engine}, \textbf{fuel}, gun, \textbf{gas}, \textbf{heat}, \textbf{cool}, \textbf{efficiency}, use, \textbf{steel}, \textbf{hot} \\
     & \textbf{file}, \textbf{character}, \textbf{print}, book, \textbf{write}, \textbf{code}, \textbf{web}, use, \textbf{page}, create\\
     & \textbf{write}, \textbf{work}, later, friend, \textbf{novel}, \textbf{publish}, life, \textbf{writing}, attend, stay \\ \cline{2-2}

    \multirow{3}{*}{\textbf{ProdLDA-Neg}} & \textbf{engine}, \textbf{fuel}, \textbf{heat}, \textbf{efficiency}, \textbf{battery}, \textbf{cool}, \textbf{hot}, \textbf{temperature}, \textbf{motor}, \textbf{machine} \\
     & \textbf{bit}, \textbf{address}, \textbf{memory}, \textbf{computer}, \textbf{implement}, \textbf{software}, \textbf{code}, \textbf{system}, \textbf{execute}, \textbf{datum} \\
     & \textbf{write}, \textbf{poem}, \textbf{poet}, \textbf{novel}, \textbf{work}, \textbf{writer}, \textbf{literary}, \textbf{poetry}, \textbf{literature}, \textbf{publish} \\ \midrule 

    \multirow{3}{*}{\textbf{NeuralLDA}\cite{srivastava2017autoencoding}} & \textbf{vehicle}, \textbf{aircraft}, capable, \textbf{missile}, \textbf{crew}, \textbf{flight}, \textbf{pilot}, car, \textbf{engine}, \textbf{weapon} \\
    & enable, \textbf{implement}, require, \textbf{computer}, \textbf{integrate}, improvement, design, \textbf{distribute}, \textbf{control}, \textbf{bit} \\
    & \textbf{work}, \textbf{write}, paris, begin, say, new, follow, \textbf{harry}, early, later \\ \cline{2-2}

    \multirow{3}{*}{\textbf{NeuralLDA-Neg}} & \textbf{aircraft}, \textbf{flight}, \textbf{airline}, \textbf{fly}, \textbf{pilot}, \textbf{missile}, \textbf{crew}, \textbf{launch}, \textbf{satellite}, \textbf{mission} \\
    & \textbf{bit}, \textbf{digital}, \textbf{computer}, \textbf{call}, \textbf{signal}, \textbf{device}, \textbf{code}, allow, provide, \textbf{application} \\
    & \textbf{publication}, \textbf{publish}, \textbf{writer}, \textbf{literary}, \textbf{magazine}, \textbf{edition}, \textbf{editor}, \textbf{theme}, \textbf{novel}, \textbf{volume} \\ \midrule 

    \multirow{3}{*}{\textbf{ETM}\cite{dieng2020topic}} & \textcolor{blue}{\textbf{film}}, \textcolor{teal}{\textbf{release}}, \textcolor{blue}{\textbf{award}}, \textcolor{blue}{\textbf{star}}, \textcolor{red}{\textbf{album}}, \textcolor{blue}{\textbf{role}}, \textcolor{red}{\textbf{song}}, million, \textcolor{blue}{\textbf{play}}, good \\
    & day, year, woman, \textbf{state}, \textbf{national}, say, \textbf{issue}, \textbf{support}, \textbf{war}, visit \\
    & \textbf{work}, \textbf{book}, \textbf{write}, life, idea, \textbf{publish}, school, man, american, new \\ \cline{2-2}

    \multirow{4}{*}{\textbf{ETM-Neg}} & \textcolor{blue}{\textbf{film}}, \textcolor{blue}{\textbf{star}}, \textcolor{blue}{\textbf{role}}, \textcolor{teal}{\textbf{release}}, \textcolor{blue}{\textbf{award}}, million, \textcolor{blue}{\textbf{play}}, \textcolor{blue}{\textbf{movie}}, \textcolor{blue}{\textbf{actor}}, good \\
    &  \textcolor{red}{\textbf{album}}, \textcolor{teal}{\textbf{release}}, \textcolor{red}{\textbf{song}}, \textcolor{red}{\textbf{music}}, \textcolor{red}{\textbf{band}}, \textcolor{red}{\textbf{single}}, \textcolor{red}{\textbf{series}}, new, \textcolor{red}{\textbf{record}}, tour \\
    & \textbf{force}, \textbf{government}, \textbf{attack}, \textbf{state}, \textbf{military}, say, \textbf{support}, \textbf{united}, \textbf{war}, \textbf{president} \\
    & \textbf{write}, \textbf{poetry}, \textbf{book}, \textbf{poet}, \textbf{poem}, \textbf{novel}, \textbf{publish}, \textbf{literary}, \textbf{story}, \textbf{translation} \\ \midrule 

    \multirow{3}{*}{\textbf{SCHOLAR}\cite{card2018neural}} & bank, \textbf{rail}, \textbf{railway}, \textbf{electricity}, port, \textbf{infrastructure}, \textbf{project}, \textbf{billion}, \textbf{investment}, customer \\
    & \textbf{physics}, \textbf{laboratory}, \textbf{professor}, \textbf{chemistry}, \textbf{mathematic}, mission, \textbf{prize}, \textbf{medal}, \textbf{space}, \textbf{fellow} \\
    & \textbf{blood}, \textbf{tissue}, \textbf{pressure}, \textbf{fluid}, \textbf{liquid}, \textbf{muscle}, \textbf{oxygen}, \textbf{radiation}, \textbf{concentration}, \textbf{molecule} \\ \cline{2-2}

    \multirow{3}{*}{\textbf{SCHOLAR-Neg}} & \textbf{rail}, \textbf{passenger}, \textbf{cable}, \textbf{railway}, \textbf{satellite}, \textbf{train}, \textbf{terminal}, \textbf{transmission}, \textbf{electricity}, \textbf{station} \\
    & \textbf{particle}, \textbf{quantum}, \textbf{electron}, \textbf{radiation}, \textbf{atom}, \textbf{galaxy}, \textbf{energy}, \textbf{universe}, \textbf{wave}, \textbf{physics} \\
    & \textbf{tissue}, \textbf{blood}, \textbf{radiation}, \textbf{patient}, \textbf{muscle}, \textbf{image}, \textbf{therapy}, \textbf{bone}, \textbf{skin}, \textbf{vision} \\ \midrule 

    \multirow{3}{*}{\textbf{GSM}\cite{miao2017discovering}} & \textbf{gun}, class, \textbf{target}, type, small, build, \textbf{war}, design, royal, large \\
    & \textbf{black}, milk, secret, house, mother, \textbf{model}, family, estate, return, price \\
    & \textbf{express}, \textbf{train}, \textbf{coach}, \textbf{railway}, \textbf{speed}, \textbf{line}, \textbf{run}, \textbf{station}, new, \textbf{service} \\ \cline{2-2}
    \multirow{3}{*}{\textbf{GSM-Neg}} & \textbf{enemy}, \textbf{strategy}, \textbf{war}, \textbf{force}, cover, \textbf{military}, system, \textbf{fire}, art, \textbf{battle} \\
    & \textbf{black}, \textbf{quantum}, \textbf{hole}, event, \textbf{theory}, \textbf{model}, \textbf{force}, \textbf{state}, \textbf{particle}, \textbf{string} \\
    & \textbf{station}, \textbf{bus}, \textbf{rail}, \textbf{rate}, \textbf{railway}, \textbf{population}, \textbf{speed}, \textbf{city}, \textbf{passenger}, \textbf{service} \\ \bottomrule
\end{tabular}
\end{adjustbox}
\end{table*}

\begin{table*}[!htbp]
\caption{Aligned topics from CLNTM, NQTM, and BAT, trained on Wiki40B for $T=120$ topics. Closely related words in each topic are highlighted in \textbf{bold}. \label{tab:Qualitative_spcls}}
\centering
\begin{adjustbox}{width=\linewidth}
  \begin{tabular}{l | l} \toprule
    \textbf{Model} & \multicolumn{1}{c}{\textbf{Topics}} \\ \midrule
    \multirow{3}{*}{\textbf{CLNTM}\cite{nguyen2021contrastive}} & \textbf{currency}, \textbf{bank}, \textbf{price}, \textbf{payment}, \textbf{financial}, \textbf{money}, \textbf{stock}, \textbf{market}, \textbf{exchange}, \textbf{investment} \\
     & \textbf{missile}, \textbf{aircraft}, \textbf{flight}, \textbf{engine}, \textbf{mission}, \textbf{launch}, \textbf{weapon}, \textbf{pilot}, \textbf{defence}, \textbf{orbit} \\
    & yoga, \textbf{novel}, \textbf{poem}, \textbf{bengali}, \textbf{poetry}, \textbf{literary}, spiritual, \textbf{publish}, sri, guru \\ \cline{2-2}

    \multirow{3}{*}{\textbf{NQTM}\cite{wu2020nqtm}} & \textbf{return}, \textbf{pay}, discover, \textbf{money}, white, president, \textbf{credit}, school, half, february \\
    & west, process, style, empire, wall, region, burn, \textbf{satellite}, classic, \textbf{moon} \\
    & \textbf{compose}, kapoor, \textbf{poet}, body, look, \textbf{poetry}, surround, farm, {character}, design \\ \cline{2-2}

    \multirow{3}{*}{\textbf{BAT}\cite{wang2020bat}} & \textbf{bank}, \textbf{credit}, \textbf{currency}, exchange, \textbf{financial}, \textbf{account}, number, contract, \textbf{business}, \textbf{finance} \\
    & \textbf{flight}, \textbf{airline}, \textbf{terminal}, \textbf{aircraft}, \textbf{air}, company, \textbf{passenger}, design, \textbf{fly}, \textbf{operate} \\
    & \textbf{award}, \textbf{hindi}, \textbf{poetry}, \textbf{work}, \textbf{write}, \textbf{poet}, collection, life, \textbf{publish}, indian \\ \bottomrule

\end{tabular}
\end{adjustbox}
\end{table*}

\subsubsection{Results} In Table \ref{tab:Qualitative}, we display the topics generated by two distinct categories of models: vanilla VAE-NTMs and their counterparts that have been enhanced with negative sampling in the decoder. We choose three topics per model. Our objective has been to analyze the effectiveness of the topic models in generating meaningful topics. To facilitate this analysis, we highlight in \textbf{bold} the words that are most semantically connected to each other within the topic, and therefore, contribute significantly to the coherence of the generated topic.

Our observations reveal that the models utilizing negative sampling consistently exhibit a higher number of these highlighted, contextually relevant words within each topic. This suggests that the incorporation of negative sampling techniques enhances the models' capacity to produce topics that are more semantically meaningful and coherent compared to those in their vanilla counterparts.

Now, let us shift our focus to an interesting observation in the case of ETM and ETM-Neg. In Table \ref{tab:Qualitative}, we have shown 3 topics for ETM and 4 topics for ETM-Neg where the first two topics in ETM-Neg are very close to the first topic in ETM. The second and third topics in ETM correspond to the third and fourth topics in ETM-Neg. The blue-highlighted words in the first ETM topic pertain to ``\textit{\textcolor{blue}{Film}}'' and the red-highlighted words in ETM relate to  ``\textit{\textcolor{red}{Music}}''. Interestingly, the word ``\textcolor{teal}{release}'' appears relevant to both themes. 
However, in the case of ETM-Neg, the first topic exclusively represents ``\textit{\textcolor{blue}{Film}}'' while the second is dedicated to ``\textit{\textcolor{red}{Music}}''. ETM fails to capture this differentiation. 

In Table \ref{tab:Qualitative_spcls}, we have presented topics generated by three different models that have no counterparts without negative sampling: CLNTM, NQTM, and BAT. The topics shown here are aligned with the topics from CombinedTM-Neg as presented in Table \ref{tab:Qualitative}. These topics cover a range of subjects, including ``Finance'' and ``Literature''. 
Notably, when we compare the quality of topics generated by these models to those in CombinedTM-Neg, we find CLNTM exhibits a relatively similar number of highlighted words for each topic when compared to CombinedTM-Neg -- this agrees with the closeness of NPMI  of CombinedTM-Neg and that of CLNTM in Table \ref{tab:QuantACcompare}. However, this consistency is not observed in the outputs of the other two models. Among them, NQTM stands out for generating topics with a higher proportion of miscellaneous words, which implies a lower level of topic cohesiveness and relevance in comparison to the benchmark set by CombinedTM-Neg.

\begin{table}[!ht]
\caption{Classification accuracy for 20NG and M10 with $T=20$ and $T=10$ topics.  Bold highlights the highest scores between models with added decoder negative sampling and their vanilla counterparts. The highest-scoring models for each dataset are marked with $\dag$. \label{tab:DocClass}}
\centering
\begin{adjustbox}{width=.8\linewidth}
    \begin{tabular}{l | c | c}
    \toprule  
     \textbf{Model} & \textbf{20NG} & \textbf{M10} \\ \midrule  
      \textbf{CombinedTM} \cite{bianchi2020pre} & 0.416 & 0.621 \\ 
      \textbf{CombinedTM-Neg} \cite{adhya2022neg} & \textbf{0.420} & \textbf{0.626} \\ \midrule  

      \textbf{ZeroShotTM} \cite{bianchi2021cross} & 0.423 & 0.648 \\ 
      \textbf{ZeroShotTM-Neg} & \textbf{0.430} & \textbf{0.654}$\dag$ \\ \midrule

      \textbf{ProdLDA} \cite{srivastava2017autoencoding} & 0.397 & 0.595 \\ 
      \textbf{ProdLDA-Neg} & \textbf{0.398} & \textbf{0.611} \\ \midrule

      \textbf{NeuralLDA} \cite{srivastava2017autoencoding} & 0.340 & 0.394 \\ 
      \textbf{NeuralLDA-Neg} & \textbf{0.345} & \textbf{0.457} \\ \midrule

      \textbf{ETM} \cite{dieng2020topic} & 0.376 & 0.408 \\
      \textbf{ETM-Neg} & \textbf{0.395} & \textbf{0.430} \\ \midrule

      \textbf{SCHOLAR} \cite{card2018neural} & \textbf{0.501} & 0.599 \\
      \textbf{SCHOLAR-Neg} & 0.499 & \textbf{0.604} \\ \midrule

      \textbf{GSM} \cite{miao2017discovering} & \textbf{0.505}$\dag$ & 0.633 \\
      \textbf{GSM-Neg} & \textbf{0.505}$\dag$ & \textbf{0.637} \\ \midrule
           
      \textbf{CLNTM} \cite{nguyen2021contrastive} & 0.439 & 0.541 \\ 
      
      \textbf{NQTM} \cite{wu2020nqtm} & 0.241 & 0.227 \\ 
      
      \textbf{BAT} \cite{wang2020bat} & 0.224 & 0.287 \\ \bottomrule  
      
    \end{tabular}
\end{adjustbox}
\end{table}

\subsection{Extrinsic Evaluation}
In this assessment, we measure the predictive performance of the generated topics by employing them in a document classification task. To conduct this evaluation, we have utilized two datasets: 20NG and M10, both sourced from OCTIS.

\subsubsection{Experimental setup}
The 20NG dataset consists of documents that are assigned to one of 20 different categories, representing their respective document labels. These categories encompass a diverse range of subjects. The M10 dataset consists of scientific papers, each labeled with one of ten distinct research areas. By evaluating how well the generated topics can predict the pre-defined class labels, we can assess the robustness and generalizability of our topic models across different domains of knowledge.

Each dataset has been partitioned into two subsets, namely the training subset and the test subset, following a ratio of 85:15. Each topic model is trained with the objective of generating a set of $T=20$ topics for the 20NG dataset and $T=10$ topics for the M10 dataset. These topics serve as latent representations of documents, expressed as $T$-dimensional topic vectors. For document classification, a linear Support Vector Machine (SVM) is employed. For every topic model, an SVM is trained with the learned representations of the training documents. The accuracy of the trained SVM models is then assessed on the test subset. 

\subsubsection{Results}
The results presented in Table \ref{tab:DocClass} indicate that, in general, models employing decoder negative sampling consistently outperform their vanilla counterparts as well as the other topic models that use contrastive learning. For the 20NG dataset, SCHOLAR's accuracy score is slightly higher than that of SCHOLAR-Neg but the difference is almost negligible. 
 
\subsection{Latent Space Visualization}
The negative sampling strategies in topic modeling aim to disentangle latent representations for dissimilar documents while enhancing similarity among representations for similar documents. To assess this capability of the topic models, we have visualized their generated latent space.

\begin{figure*}
    \centering
    \includegraphics[width=\linewidth]{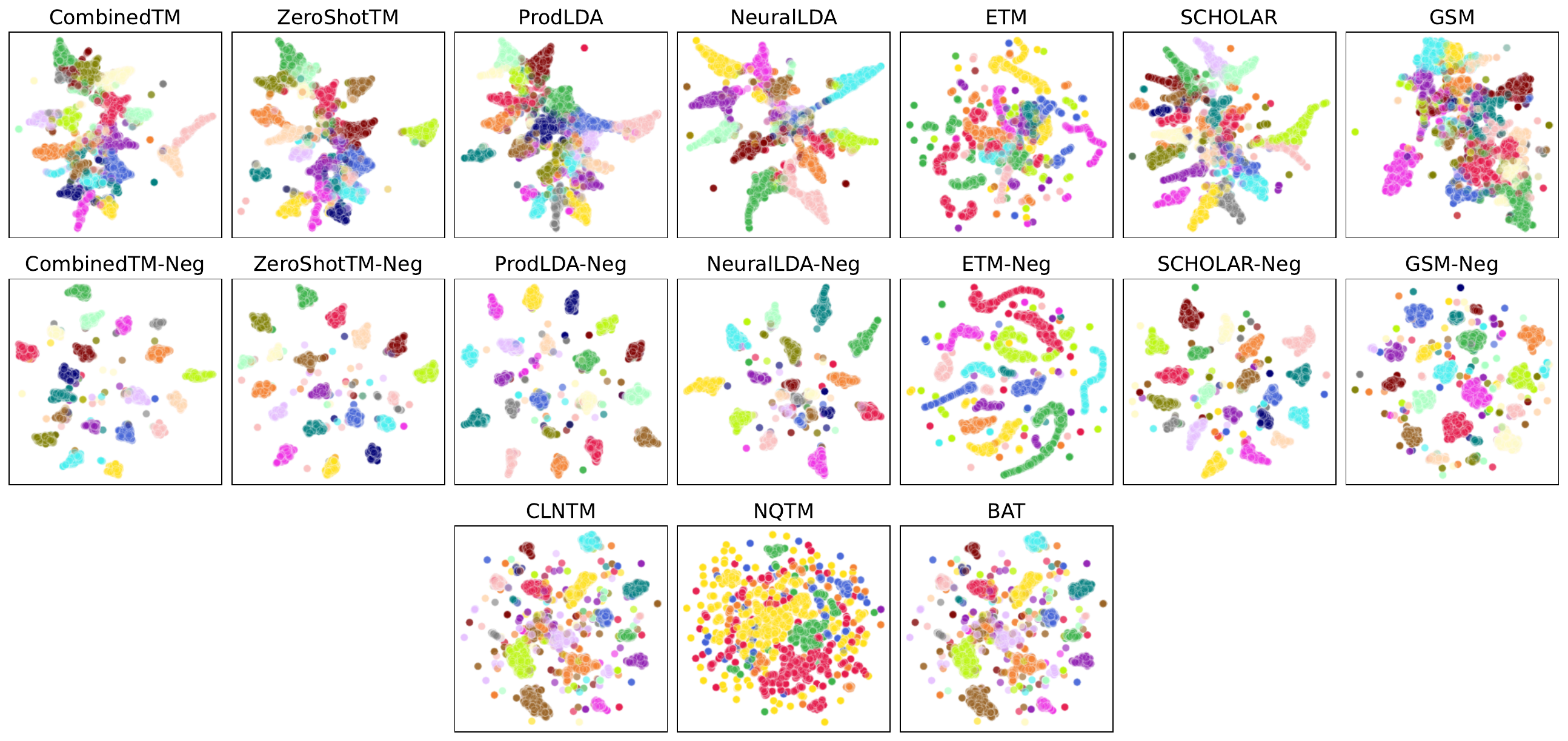}
    \caption{Latent space visualization is shown for different topic models on the 20NG dataset for a topic count of 20.} \label{fig:CompareLatentViz}
\end{figure*}

\subsubsection{Experimental setup} 
We visualized the latent space using the 20NG dataset, training each model for a topic count of 20. After training, each document is labeled with the topic that has the highest proportion within it. We then employed \textit{Uniform Manifold Approximation and Projection} (UMAP) for robust dimensionality reduction \cite{mcinnes2020umap}. UMAP transformed the document-topic distributions into two-dimensional representations for visualization. 

\subsubsection{Results} 
In Figure \ref{fig:CompareLatentViz}, we illustrate the disentanglement of the 20 distinct clusters within the 20NG dataset across all topic models. The results demonstrate that, following the application of negative sampling in the decoder, document clusters exhibit clear separation, a characteristic not observed in vanilla VAE-NTMs. Our proposed method also shows better cluster separation compared to that produced by other negative sampling-based models, namely, CLNTM, NQTM, and BAT. 

\section{Robustness Assessment of Proposed Methodology \label{sec:robustness}}

We analyzed the impact of increasing vocabulary size on our proposed methodology. Furthermore, we explored whether the proposed negative sampling methodology results in an increase in the number of trainable parameters or model training time. We also investigated the trade-off between performance improvement and these complexity changes.
\subsection{Impact of Vocabulary Size}
In this analysis, we study how increasing the vocabulary size of the corpus affects the model performance.

\subsubsection{Experimental Setup}
Among the four datasets, the 20NG and M10 datasets are available in OCTIS in pre-processed formats, thus having fixed vocabulary sizes. Therefore, for this experiment, we focused on the Wiki40B dataset and varied the vocabulary size within the range of $\{ 1000, 2000, 4000, 8000\}$. We conducted experiments by running all VAE-NTMs and their negative sampling counterparts across these vocabulary sizes while keeping the topic count fixed at 120 for five different random runs.

\begin{figure}
    \centering
    \includegraphics[width=\linewidth]{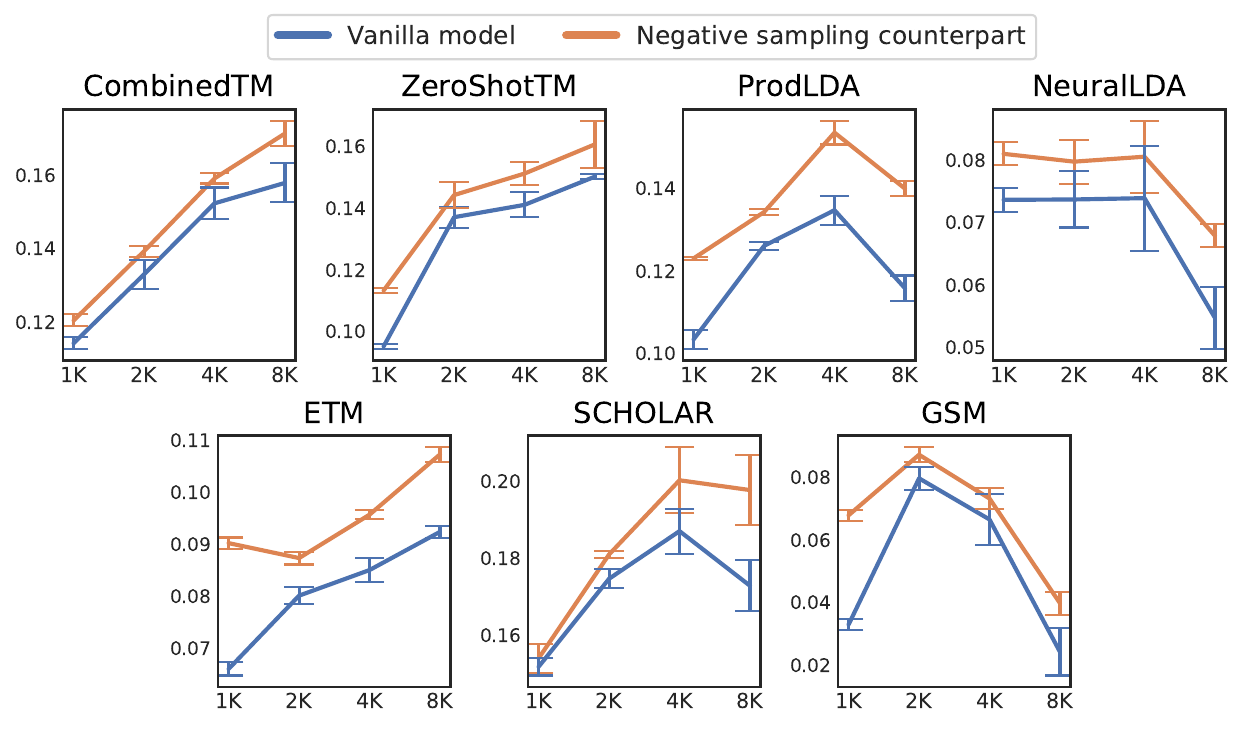}
    \caption{The mean and variance of the NPMI scores on the Wiki40B dataset are shown for a topic count of 120 across all VAE-NTMs and their negative sampling-based counterparts, varying the vocabulary size in $\{1000, 2000, 4000, 8000\}$.} \label{fig:Ablation_Vsize}
\end{figure}

\subsubsection{Results} The results are visualized in Figure \ref{fig:Ablation_Vsize}, with each subplot displaying the outcomes of a specific model and its decoder negative sampling-based counterpart. The X-axis, shown in the logarithmic scale, represents the vocabulary size, while the Y-axis represents the NPMI scores of the models. Analysis of the figure suggests that the models with negative sampling strategy always perform better than their vanilla counterparts for a given vocabulary size. In some cases, the NPMI steadily increases with vocabulary size, and in other cases, it shows a more complicated pattern; however, the behavior is similar for the baseline and the corresponding negative sampling-based models.

\begin{table*}[!htbp]
\caption{Training time versus performance trade-off is analyzed using the 20NG dataset with 20 topics. Mean and variance values are shown for both training time and NPMI scores, along with the percentage increase in time and NPMI (using their mean values).
\label{tab:Complexity_Anal}}
\centering
\begin{adjustbox}{width=.75\linewidth}
  \begin{tabular}{l|cc|cc} \toprule 
\textbf{Model} & \textbf{Time (sec.)} & \textbf{Inc. (\%)} & \textbf{NPMI} & \textbf{Inc. (\%)}\\ \midrule 

\textbf{CombinedTM} \cite{bianchi2020pre} & $48.4 \pm 0.3$ & \multirow{2}{*}{34.2\%} & $0.107 \pm 0.004$ & \multirow{2}{*}{14.02\%}\\ 
\textbf{CombinedTM-Neg} \cite{adhya2022neg} & $65.0 \pm 0.7$ &                       & $0.122 \pm 0.001$ & \\ \midrule

\textbf{ZeroShotTM} \cite{bianchi2021cross} & $43.5 \pm 1.0$ & \multirow{2}{*}{35.6\%}  & $0.113 \pm 0.009$ & \multirow{2}{*}{11.50\%}\\ 
\textbf{ZeroShotTM-Neg}                     & $59.0 \pm 1.0$ &                          & $0.126 \pm 0.002$ & \\ \midrule

\textbf{ProdLDA} \cite{srivastava2017autoencoding} & $34.6 \pm 1.2$ & \multirow{2}{*}{48.0\%} & $0.086 \pm 0.004$ & \multirow{2}{*}{20.93\%}\\
\textbf{ProdLDA-Neg}                               & $51.2 \pm 0.9$ &                         & $0.104 \pm 0.004$ & \\ \midrule

\textbf{NeuralLDA} \cite{srivastava2017autoencoding} & $35.7 \pm 0.1$ & \multirow{2}{*}{42.6\%} & $0.051 \pm 0.005$ & \multirow{2}{*}{35.29\%}\\
\textbf{NeuralLDA-Neg}                               & $50.9 \pm 1.6$ &                         & $0.069 \pm 0.006$ & \\ \midrule

\textbf{ETM} \cite{dieng2020topic} & $30.4 \pm 0.4$ & \multirow{2}{*}{33.6\%} & $0.055 \pm 0.003$ & \multirow{2}{*}{10.91\%}\\
\textbf{ETM-Neg}                   & $40.6 \pm 0.9$ &                         & $0.061 \pm 0.003$ \\ \midrule

\textbf{SCHOLAR} \cite{hu2020neural} & $121.3 \pm 13.0$ & \multirow{2}{*}{48.9\%} & $0.171 \pm 0.003$ & \multirow{2}{*}{6.43\%}\\
\textbf{SCHOLAR-Neg}                 & $180.6 \pm 44.5$ &                         & $0.182 \pm 0.003$ \\ \midrule

\textbf{GSM} \cite{miao2017discovering} & $72.6\pm 0.4$ & \multirow{2}{*}{33.9\%} & $0.093 \pm 0.004$ & \multirow{2}{*}{9.68\%}\\
\textbf{GSM-Neg}                        & $97.2 \pm 0.6$ &                        & $0.102 \pm 0.005$ \\ \bottomrule

\end{tabular}
\end{adjustbox}
\end{table*}

\subsection{Training Time Analysis}
The trainable parameter count remains unchanged when negative sampling is incorporated into a VAE-NTM. This is because the negative sampling algorithm does not introduce any additional trainable parameters, as clearly visible in Algorithm \ref{alg:CombinedTM-Neg}. However, the training time of the model increases due to the computation of negative samples and the triplet loss term. We have examined the trade-off between training time and performance, as measured by the NPMI value, across all VAE-NTMs and their respective negative sampling-based counterparts.

\subsubsection{Experimental setup}
We trained all models using the 20NG dataset with a topic count of 20 for five distinct random seeds. We then measured their run times and computed the topic coherence scores (NPMI values). The results are presented in Table \ref{tab:Complexity_Anal}.

\subsubsection{Results}
As shown in Table \ref{tab:Complexity_Anal}, the inclusion of the negative sampling strategy in the VAE-NTMs leads to an increase in the NPMI score ranging from a minimum of 6.43\% to a maximum of 35.29\%. However, this performance improvement comes at a cost in terms of time complexity. The training time increases, ranging from 33.6\% to 48.9\%.

\begin{table*}[!htbp]
\caption{Topics from GPTopic trained on 20NG for $T=20$ topics. Closely related words in each topic are highlighted in \textbf{bold}. \label{tab:Qualitative_GPTopic}}
\centering
\begin{adjustbox}{width=\linewidth}
  \begin{tabular}{l | l} \toprule
    \textbf{Model} & \multicolumn{1}{c}{\textbf{Topics}} \\ \midrule
    
    \multirow{4}{*}{\textbf{GPTopic}\cite{reuter2024gptopic}} & \textbf{armenian}, \textbf{turkish}, \textbf{israeli}, \textbf{jewish}, account, argue, \textbf{genocide}, \textbf{attack}, \textbf{arab}, \textbf{greek} \\
     & \textbf{encryption}, \textbf{clipper}, \textbf{phone}, \textbf{escrow}, wait, \textbf{security}, ticket, \textbf{secure}, stock, agency \\
    & \textbf{monitor}, \textbf{modem}, kind, \textbf{port}, \textbf{vga}, learn, \textbf{mhz}, mode, height, \textbf{serial} \\
    & \textbf{player}, profit, \textbf{hit}, guilty, \textbf{goal}, motor, \textbf{baseball}, illegal, \textbf{playoff}, \textbf{period} \\ \midrule

    \multirow{4}{*}{\textbf{CombinedTM-Neg}\cite{adhya2022neg}} & \textbf{turkish}, \textbf{armenian}, \textbf{jewish}, \textbf{population}, \textbf{muslim}, \textbf{village}, \textbf{israeli}, \textbf{genocide}, \textbf{government}, \textbf{war} \\
        & \textbf{chip}, \textbf{key}, \textbf{encryption}, \textbf{government}, \textbf{clipper},  \textbf{phone}, \textbf{security}, \textbf{privacy}, \textbf{escrow}, \textbf{secure} \\
        & \textbf{video}, \textbf{monitor}, \textbf{vga}, \textbf{port}, \textbf{modem}, \textbf{apple}, \textbf{driver}, \textbf{card}, \textbf{resolution}, \textbf{board}  \\ 
        & \textbf{score}, \textbf{playoff}, \textbf{period}, \textbf{play}, \textbf{fan}, \textbf{win}, \textbf{hockey}, \textbf{game}, \textbf{baseball}, \textbf{lose} \\
        \bottomrule
        
\end{tabular}
\end{adjustbox}
\end{table*}

\section{Comparison with LLM-based Topic Models \label{sec:LLM_based}}
Large language model (LLM)-based topic models \cite{wang2023prompting, pham2024topicgpt, reuter2024gptopic} have recently emerged as tools for modeling user interactions and guiding topic generation based on needs and domain knowledge. A user can prompt LLMs to extract topics from documents. This flexible prompt-based interaction enables non-technical users to use topic models easily.
However, they exhibit significant differences from classical or neural topic models, which function as statistical models for document generation. Traditional models operate under the assumption that a document is a distribution over topics, with each topic being a distribution over words. Various inference techniques are used to learn the distribution parameters. In contrast, LLM-based topic models rely on prompt engineering and generate topics as labels where a label is typically a word or a short phrase rather than a set of top-ranked words. They do not output any statistical distribution to quantify the topics and their presence in the documents. However, using an LLM to extract topics from a large collection of documents is computationally expensive \cite{wang2023prompting, pham2024topicgpt}. 
To mitigate this, a sample of documents has to be intelligently selected for input to the model. However, this might reduce the output topic quality compared to traditional topic models \cite{pham2024topicgpt}. Additional issues include limits on the context length and the closed-source nature of many high-performance LLMs. In industrial applications, employing external LLM services for topic modeling on sensitive data can potentially lead to data leakage.

We have conducted experiments on the 20NG dataset with GPTopic \cite{reuter2024gptopic} which leverages the \texttt{gpt-3.5-turbo-16k}\footnote{\url{https://platform.openai.com/docs/models/gpt-3-5-turbo}} model. The topic count was set to 20. The scores for (NPMI, CV, IRBO) are (0.113, 0.607, 0.985). In contrast, our proposed decoder negative sampling-based topic models produce the following results: CombinedTM-Neg (0.129, 0.672, 0.994), ZeroShotTM-Neg (0.135, 0.679, 0.996), SCHOLAR-Neg (0.193, 0.801, 0.999), and ProdLDA-Neg (0.115, 0.656, 0.992). We observe our models have generated topics with better coherence and diversity scores than those of GPTopic.

For further evaluation, in Table \ref{tab:Qualitative_GPTopic}, we showcase the top 10 words of 4 selected topics generated by GPTopic \cite{reuter2024gptopic} on the 20NG dataset for a topic count of 20, alongside the corresponding topics generated by CombinedTM-Neg \cite{adhya2022neg}. GPTopic captures topics similar to CombinedTM-Neg, but some of the topics from GPTopic include out-of-domain words that appear to be incoherent with the remaining words in the respective topics.

\section{Conclusion and Future Work} \label{sec:conclusion}
Our study presents a comparative analysis of negative sampling-based neural topic models. We have evaluated their performance through several different evaluation procedures, including quantitative review, qualitative assessment, extrinsic evaluation concerning document classification tasks, and visualization of latent topic spaces. To execute this investigation, we have made use of four publicly available datasets. The results obtained from all these experiments highlight that incorporating negative sampling into the decoder of the VAE in a neural topic model leads to improved performance in terms of higher topic coherence, larger topic diversity, and enhanced accuracy in document classification. 
Thus, this approach holds significant promise for enhancing topic models. In the future, we aim to seek a theoretical explanation for the superior efficacy of the negative sampling approach within the decoder of VAE-based neural topic models, as compared to its counterpart in the encoder.


\begin{thebibliography}{10}
\providecommand{\url}[1]{#1}
\csname url@samestyle\endcsname
\providecommand{\newblock}{\relax}
\providecommand{\bibinfo}[2]{#2}
\providecommand{\BIBentrySTDinterwordspacing}{\spaceskip=0pt\relax}
\providecommand{\BIBentryALTinterwordstretchfactor}{4}
\providecommand{\BIBentryALTinterwordspacing}{\spaceskip=\fontdimen2\font plus
\BIBentryALTinterwordstretchfactor\fontdimen3\font minus \fontdimen4\font\relax}
\providecommand{\BIBforeignlanguage}[2]{{%
\expandafter\ifx\csname l@#1\endcsname\relax
\typeout{** WARNING: IEEEtran.bst: No hyphenation pattern has been}%
\typeout{** loaded for the language `#1'. Using the pattern for}%
\typeout{** the default language instead.}%
\else
\language=\csname l@#1\endcsname
\fi
#2}}
\providecommand{\BIBdecl}{\relax}
\BIBdecl

\bibitem{blei2017variational}
D.~M. Blei, A.~Kucukelbir, and J.~D. McAuliffe, ``Variational inference: A review for statisticians,'' \emph{Journal of the American Statistical Association}, vol. 112, no. 518, pp. 859--877, 2017.

\bibitem{miao2016neural}
Y.~Miao, L.~Yu, and P.~Blunsom, ``Neural variational inference for text processing,'' in \emph{Proceedings of the 33rd International Conference on Machine Learning}, vol.~48.\hskip 1em plus 0.5em minus 0.4em\relax PMLR, 20--22 Jun 2016, pp. 1727--1736.

\bibitem{miao2017discovering}
Y.~Miao, E.~Grefenstette, and P.~Blunsom, ``Discovering discrete latent topics with neural variational inference,'' in \emph{Proceedings of the 34th International Conference on Machine Learning}, vol.~70.\hskip 1em plus 0.5em minus 0.4em\relax PMLR, Aug. 2017, pp. 2410--2419.

\bibitem{srivastava2017autoencoding}
A.~Srivastava and C.~Sutton, ``Autoencoding variational inference for topic models,'' in \emph{International Conference on Learning Representations}, 2017.

\bibitem{card2018neural}
D.~Card, C.~Tan, and N.~A. Smith, ``Neural models for documents with metadata,'' in \emph{Proceedings of the 56th Annual Meeting of the Association for Computational Linguistics (Volume 1: Long Papers)}.\hskip 1em plus 0.5em minus 0.4em\relax Association for Computational Linguistics, Jul. 2018, pp. 2031--2040.

\bibitem{dieng2020topic}
A.~B. Dieng, F.~J.~R. Ruiz, and D.~M. Blei, ``Topic modeling in embedding spaces,'' \emph{Transactions of the Association for Computational Linguistics}, vol.~8, pp. 439--453, 2020.

\bibitem{bianchi2020pre}
F.~Bianchi, S.~Terragni, and D.~Hovy, ``Pre-training is a hot topic: Contextualized document embeddings improve topic coherence,'' in \emph{Proceedings of the 59th Annual Meeting of the Association for Computational Linguistics and the 11th International Joint Conference on Natural Language Processing (Volume 2: Short Papers)}.\hskip 1em plus 0.5em minus 0.4em\relax Association for Computational Linguistics, Aug. 2021, pp. 759--766.

\bibitem{bianchi2021cross}
F.~Bianchi, S.~Terragni, D.~Hovy, D.~Nozza, and E.~Fersini, ``Cross-lingual contextualized topic models with zero-shot learning,'' in \emph{Proceedings of the 16th Conference of the European Chapter of the Association for Computational Linguistics: Main Volume}.\hskip 1em plus 0.5em minus 0.4em\relax Association for Computational Linguistics, Apr. 2021, pp. 1676--1683.

\bibitem{adhya2024ginopic}
S.~Adhya and D.~K. Sanyal, ``{GIN}opic: Topic modeling with graph isomorphism network,'' in \emph{Proceedings of the 2024 Conference of the North American Chapter of the Association for Computational Linguistics: Human Language Technologies (Volume 1: Long Papers)}.\hskip 1em plus 0.5em minus 0.4em\relax Association for Computational Linguistics, Jun. 2024, pp. 6171--6183.

\bibitem{xu2022negative}
L.~Xu, J.~Lian, W.~X. Zhao, M.~Gong, L.~Shou, D.~Jiang, X.~Xie, and J.~rong Wen, ``Negative sampling for contrastive representation learning: A review,'' \emph{arXiv preprint arXiv:2206.00212}, 2022.

\bibitem{yang2024does}
Z.~Yang, M.~Ding, T.~Huang, Y.~Cen, J.~Song, B.~Xu, Y.~Dong, and J.~Tang, ``Does negative sampling matter? a review with insights into its theory and applications,'' \emph{IEEE Transactions on Pattern Analysis and Machine Intelligence}, vol.~46, no.~8, pp. 5692--5711, 2024.

\bibitem{adhya2022neg}
S.~Adhya, A.~Lahiri, D.~Kumar~Sanyal, and P.~Pratim~Das, ``Improving contextualized topic models with negative sampling,'' in \emph{Proceedings of the 19th International Conference on Natural Language Processing (ICON)}.\hskip 1em plus 0.5em minus 0.4em\relax Association for Computational Linguistics, Dec. 2022, pp. 128--138.

\bibitem{blei2003lda}
D.~M. Blei, A.~Y. Ng, and M.~I. Jordan, ``Latent {D}irichlet {A}llocation,'' \emph{Journal of Machine Learning Research}, vol.~3, no. Jan., pp. 993--1022, 2003.

\bibitem{gelfand1990mcmc}
A.~E. Gelfand and A.~F.~M. Smith, ``Sampling-based approaches to calculating marginal densities,'' \emph{Journal of the American Statistical Association}, vol.~85, no. 410, pp. 398--409, 1990.

\bibitem{minka2002expectation}
T.~Minka and J.~Lafferty, ``Expectation-propagation for the generative aspect model,'' in \emph{Proceedings of the 18th Conference on Uncertainty in Artificial Intelligence}.\hskip 1em plus 0.5em minus 0.4em\relax Morgan Kaufmann Publishers Inc., 2002, p. 352–359.

\bibitem{kingma2013auto}
D.~P. Kingma and M.~Welling, ``Auto-encoding variational bayes,'' \emph{arXiv preprint arXiv:1312.6114}, 2013.

\bibitem{chopra2005learning}
S.~Chopra, R.~Hadsell, and Y.~LeCun, ``Learning a similarity metric discriminatively, with application to face verification,'' in \emph{Proceedings of the 2005 IEEE Computer Society Conference on Computer Vision and Pattern Recognition (CVPR'05)}, vol.~1, 2005, pp. 539--546 vol. 1.

\bibitem{mikolov2013w2v}
T.~Mikolov, I.~Sutskever, K.~Chen, G.~S. Corrado, and J.~Dean, ``Distributed representations of words and phrases and their compositionality,'' in \emph{Advances in Neural Information Processing Systems}, vol.~26.\hskip 1em plus 0.5em minus 0.4em\relax Curran Associates, Inc., 2013.

\bibitem{nguyen2021contrastive}
T.~Nguyen and A.~T. Luu, ``Contrastive learning for neural topic model,'' in \emph{Advances in Neural Information Processing Systems}, vol.~34.\hskip 1em plus 0.5em minus 0.4em\relax Curran Associates, Inc., 2021, pp. 11\,974--11\,986.

\bibitem{wu2020nqtm}
X.~Wu, C.~Li, Y.~Zhu, and Y.~Miao, ``Short text topic modeling with topic distribution quantization and negative sampling decoder,'' in \emph{Proceedings of the 2020 Conference on Empirical Methods in Natural Language Processing (EMNLP)}.\hskip 1em plus 0.5em minus 0.4em\relax Association for Computational Linguistics, Nov. 2020, pp. 1772--1782.

\bibitem{wang2018atm}
R.~Wang, D.~Zhou, and Y.~He, ``{ATM}: Adversarial-neural topic model,'' \emph{Information Processing \& Management}, vol.~56, no.~6, p. 102098, 2019.

\bibitem{wang2020bat}
R.~Wang, X.~Hu, D.~Zhou, Y.~He, Y.~Xiong, C.~Ye, and H.~Xu, ``Neural topic modeling with bidirectional adversarial training,'' in \emph{Proceedings of the 58th Annual Meeting of the Association for Computational Linguistics}.\hskip 1em plus 0.5em minus 0.4em\relax Association for Computational Linguistics, Jul. 2020, pp. 340--350.

\bibitem{hu2020neural}
X.~Hu, R.~Wang, D.~Zhou, and Y.~Xiong, ``Neural topic modeling with cycle-consistent adversarial training,'' in \emph{Proceedings of the 2020 Conference on Empirical Methods in Natural Language Processing (EMNLP)}.\hskip 1em plus 0.5em minus 0.4em\relax Association for Computational Linguistics, Nov. 2020, pp. 9018--9030.

\bibitem{terragni2020octis}
S.~Terragni, E.~Fersini, B.~G. Galuzzi, P.~Tropeano, and A.~Candelieri, ``{OCTIS}: Comparing and optimizing topic models is simple!'' in \emph{Proceedings of the 16th Conference of the European Chapter of the Association for Computational Linguistics: System Demonstrations}.\hskip 1em plus 0.5em minus 0.4em\relax Association for Computational Linguistics, Apr. 2021, pp. 263--270.

\bibitem{lisena2020tomodapi}
P.~Lisena, I.~Harrando, O.~Kandakji, and R.~Troncy, ``{TOMODAPI}: A topic modeling {API} to train, use and compare topic models,'' in \emph{Proceedings of Second Workshop for NLP Open Source Software (NLP-OSS)}.\hskip 1em plus 0.5em minus 0.4em\relax Association for Computational Linguistics, Nov. 2020, pp. 132--140.

\bibitem{mackay1998choice}
D.~J. MacKay, ``Choice of basis for {Laplace} approximation,'' \emph{Machine Learning}, vol.~33, pp. 77--86, 1998.

\bibitem{oord2019representation}
A.~van~den Oord, Y.~Li, and O.~Vinyals, ``Representation learning with contrastive predictive coding,'' \emph{arXiv preprint arXiv:1807.03748}, 2018.

\bibitem{Qiang2020Short}
J.~Qiang, Z.~Qian, Y.~Li, Y.~Yuan, and X.~Wu, ``Short text topic modeling techniques, applications, and performance: A survey,'' \emph{IEEE Transactions on Knowledge and Data Engineering}, vol.~34, no.~3, pp. 1427--1445, 2022.

\bibitem{Pan2016M10}
S.~Pan, J.~Wu, X.~Zhu, C.~Zhang, and Y.~Wang, ``Tri-party deep network representation,'' in \emph{Proceedings of the 25th International Joint Conference on Artificial Intelligence}.\hskip 1em plus 0.5em minus 0.4em\relax AAAI Press, 2016, p. 1895–1901.

\bibitem{Wiki40B}
M.~Guo, Z.~Dai, D.~Vrande{\v{c}}i{\'c}, and R.~Al-Rfou, ``\BIBforeignlanguage{English}{{W}iki-40{B}: Multilingual language model dataset},'' in \emph{\BIBforeignlanguage{English}{Proceedings of the 12th Language Resources and Evaluation Conference}}.\hskip 1em plus 0.5em minus 0.4em\relax European Language Resources Association, May 2020, pp. 2440--2452.

\bibitem{reimers2019sentence}
N.~Reimers and I.~Gurevych, ``Sentence-{BERT}: Sentence embeddings using {S}iamese {BERT}-networks,'' in \emph{Proceedings of the 2019 Conference on Empirical Methods in Natural Language Processing and the 9th International Joint Conference on Natural Language Processing (EMNLP-IJCNLP)}.\hskip 1em plus 0.5em minus 0.4em\relax Association for Computational Linguistics, Nov. 2019, pp. 3982--3992.

\bibitem{hoyle2021automated}
A.~Hoyle, P.~Goel, A.~Hian-Cheong, D.~Peskov, J.~Boyd-Graber, and P.~Resnik, ``Is automated topic model evaluation broken? the incoherence of coherence,'' in \emph{Advances in Neural Information Processing Systems}, vol.~34.\hskip 1em plus 0.5em minus 0.4em\relax Curran Associates, Inc., 2021, pp. 2018--2033.

\bibitem{lau2014machine}
J.~H. Lau, D.~Newman, and T.~Baldwin, ``Machine reading tea leaves: Automatically evaluating topic coherence and topic model quality,'' in \emph{Proceedings of the 14th Conference of the {E}uropean Chapter of the Association for Computational Linguistics}.\hskip 1em plus 0.5em minus 0.4em\relax Association for Computational Linguistics, Apr. 2014, pp. 530--539.

\bibitem{roder2015exploring}
M.~R\"{o}der, A.~Both, and A.~Hinneburg, ``Exploring the space of topic coherence measures,'' in \emph{Proceedings of the 8th ACM International Conference on Web Search and Data Mining}.\hskip 1em plus 0.5em minus 0.4em\relax Association for Computing Machinery, 2015, p. 399–408.

\bibitem{krasnashchok2018improving}
K.~Krasnashchok and S.~Jouili, ``Improving topic quality by promoting named entities in topic modeling,'' in \emph{Proceedings of the 56th Annual Meeting of the Association for Computational Linguistics (Volume 2: Short Papers)}.\hskip 1em plus 0.5em minus 0.4em\relax Melbourne, Australia: Association for Computational Linguistics, Jul. 2018, pp. 247--253.

\bibitem{webber2010similarity}
W.~Webber, A.~Moffat, and J.~Zobel, ``A similarity measure for indefinite rankings,'' \emph{ACM Transactions on Information Systems}, vol.~28, no.~4, Nov. 2010.

\bibitem{adhya2023neural}
S.~Adhya, A.~Lahiri, and D.~K. Sanyal, ``Do neural topic models really need dropout? analysis of the effect of dropout in topic modeling,'' in \emph{Proceedings of the 17th Conference of the European Chapter of the Association for Computational Linguistics}.\hskip 1em plus 0.5em minus 0.4em\relax Association for Computational Linguistics, May 2023, pp. 2220--2229.

\bibitem{mcinnes2020umap}
L.~McInnes, J.~Healy, and J.~Melville, ``{UMAP}: Uniform manifold approximation and projection for dimension reduction,'' \emph{arXiv preprint arXiv:1802.03426}, 2018.

\bibitem{reuter2024gptopic}
A.~Reuter, A.~Thielmann, C.~Weisser, S.~Fischer, and B.~S{\"a}fken, ``{GPTopic}: Dynamic and interactive topic representations,'' \emph{arXiv preprint arXiv:2403.03628}, 2024.

\bibitem{wang2023prompting}
H.~Wang, N.~Prakash, N.~Hoang, M.~Hee, U.~Naseem, and R.~Lee, ``Prompting large language models for topic modeling,'' in \emph{Proceedings of the 2023 IEEE International Conference on Big Data (BigData)}.\hskip 1em plus 0.5em minus 0.4em\relax IEEE Computer Society, dec 2023, pp. 1236--1241.

\bibitem{pham2024topicgpt}
C.~Pham, A.~Hoyle, S.~Sun, P.~Resnik, and M.~Iyyer, ``{T}opic{GPT}: A prompt-based topic modeling framework,'' in \emph{Proceedings of the 2024 Conference of the North American Chapter of the Association for Computational Linguistics: Human Language Technologies (Volume 1: Long Papers)}.\hskip 1em plus 0.5em minus 0.4em\relax Association for Computational Linguistics, Jun. 2024, pp. 2956--2984.

\end{thebibliography}

\begin{IEEEbiography}
[{\includegraphics[width=1in,height=1.25in,clip,keepaspectratio]{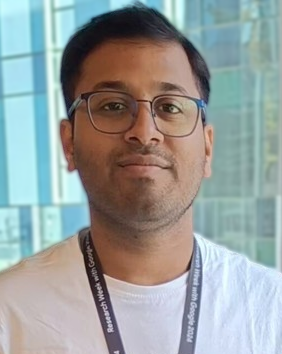}}]{Suman Adhya}{\space}received his B.Sc. (Honours) degree in Mathematics from the University of Calcutta in 2018 and M.Sc. degree in Mathematics and Computing from the Indian Association for the Cultivation of Science (IACS) in 2020. He is currently pursuing his Ph.D. in Computer Science at IACS. His research interests include topic modeling, natural language processing, and machine learning.
\end{IEEEbiography}

\begin{IEEEbiography}
[{\includegraphics[width=1in,height=1.25in,clip,keepaspectratio]{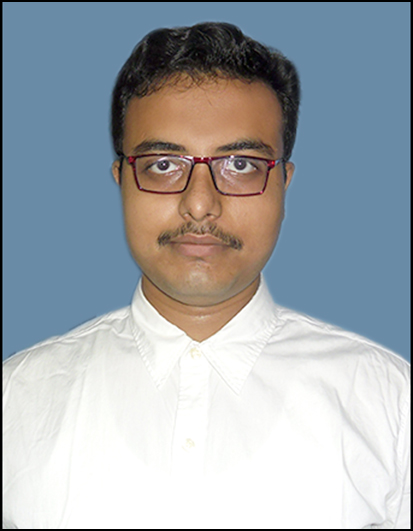}}]{Avishek Lahiri}{\space} received his B.Sc.(Honours) degree in Computer Science from the University of Calcutta in 2020 and completed his M.Sc. in Mathematics and Computing from the Indian Association for the Cultivation of Science (IACS) in 2022. Currently, he is pursuing his PhD in Computer Science at IACS. His research interests include information extraction, natural language processing, and machine learning.

\end{IEEEbiography}

\begin{IEEEbiography}[{\includegraphics[width=1in,height=1.25in,clip,keepaspectratio]{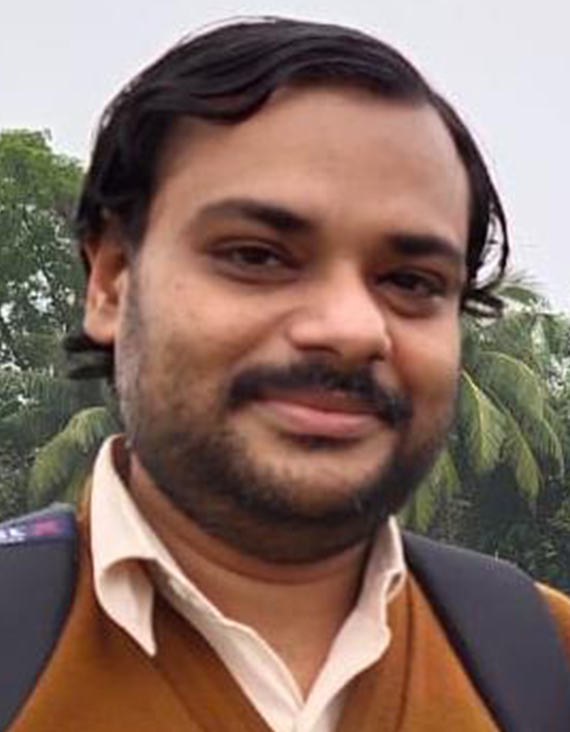}}]{Debarshi Kumar Sanyal}{\space}received his B.E. degree in Information Technology and his Ph.D. in Engineering from Jadavpur University, Kolkata, in 2005 and 2012, respectively. He currently serves as an Assistant Professor in the School of Mathematical and Computational Sciences at the Indian Association for the Cultivation of Science, Kolkata, India. His current research interests include natural language processing, digital library technologies, information retrieval, and machine learning.
\end{IEEEbiography}

\begin{IEEEbiography}[{\includegraphics[width=1in,height=1.25in,clip,keepaspectratio]{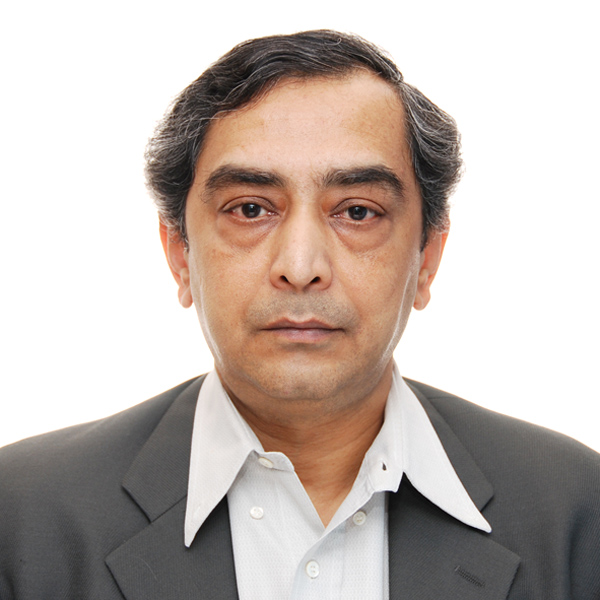}}]{Partha Pratim Das} (Member, IEEE) received the B.Tech., M.Tech., and Ph.D. degrees from the Department of Electronics and Electrical Communication, IIT Kharagpur, in 1984, 1985, and 1988 respectively. 

He is a Visiting Professor with the Department of Computer Science, Ashoka University. He is on leave from IIT Kharagpur, where he was a Professor with the Department of Computer Science and Engineering. He has over 22 years of experience in teaching and research with IIT Kharagpur, and about 13 years of experience in software industry, including start-ups. Over the past ten years, he has led the Development of National Digital Library of India (NDLI) Project, MoE, GoI, as a Joint Principal Investigator. He has also developed a unique vertical DEEPAK: Disability Education and Engagement Portal for Access to Knowledge. He has also led engineering entrepreneurship education, research, facilitation, and deployment with IIT Kharagpur, from 2013 to 2020. He is a strong proponent of online education. He has been a Key Instructor of three courses with SWAYAM-NPTEL, since 2016. During the pandemic, he was instrumental in making learning material available to the students through NDLI. He currently works on the following problems: hands-free control and immersive navigation of Chandrayaan and Mangalyaan images on large displays (with ISRO), smart knowledge transfer for legacy software projects, automated interpretation of Bharatanatyam dance, and the development of Indian food atlas and food knowledge graph. 

Dr. Das received the Young Scientist/Engineer Award from the Indian National Science Academy, in 1990, the Indian National Academy of Engineering, in 1996, and the Indian Academy of Sciences, in 1992. Being in the leadership team of NDLI, he was recognized for his contributions toward online education during the pandemic through several awards, including the OE Awards for Excellence: Open Resilience, in 2020; the SM4E Award: Innovation@COVID-19, in 2021; and the World Summit Award: Learning and Education, in 2021.
\end{IEEEbiography}

\end{document}